\documentclass[sigconf, screen]{acmart}
\usepackage{balance}
\usepackage{pifont}
\usepackage{subcaption}
\usepackage{multirow}
\usepackage{colortbl}

\usepackage{booktabs}
\usepackage{graphicx}

\begin{document}

\title{Comparison Drives Preference: Reference-Aware Modeling for AI-Generated Video Quality Assessment}

\author{Minghao Zou}
\affiliation{%
	\institution{Cardiff University}
	\city{Cardiff}
	\country{UK}
}
\email{zoum1@cardiff.ac.uk}

\author{Gen Liu}
\affiliation{%
	\institution{Shandong University of Science and Technology}
	\city{Qingdao}
	\country{China}
}
\email{lg97@sdust.edu.cn}
\authornote{Corresponding authors.}

\author{Guanghui Yue}
\affiliation{%
	\institution{Shenzhen University}
	\city{Shenzhen}
	\country{China}
}
\email{yueguanghui@szu.edu.cn}

\author{Baoquan Zhao}
\affiliation{%
	\institution{Sun Yat-sen University}
	\city{Guangzhou}
	\country{China}
}
\email{zhaobaoquan@mail.sysu.edu.cn}

\author{Zhihua Wang}
\affiliation{%
	\institution{City University of Hong Kong}
	\city{Hong Kong}
	\country{China}
}
\email{zhihua.wang@cityu.edu.hk}

\author{Paul L. Rosin}
\affiliation{%
	\institution{Cardiff University}
	\city{Cardiff}
	\country{UK}
}
\email{rosinpl@cardiff.ac.uk}

\author{Hantao Liu}
\affiliation{%
	\institution{Cardiff University}
	\city{Cardiff}
	\country{UK}
}
\email{liuh35@cardiff.ac.uk}

\author{Wei Zhou}
\affiliation{%
	\institution{Cardiff University}
	\city{Cardiff}
	\country{UK}
}
\email{zhouw26@cardiff.ac.uk}
\authornotemark[1]
\renewcommand{\shortauthors}{Minghao Zou et al.}

\begin{abstract}
	The rapid advancement of generative models has led to a growing volume of AI-generated videos, making the automatic quality assessment of such videos increasingly important. Existing AI-generated content video quality assessment (AIGC-VQA) methods typically estimate visual quality by analyzing each video independently, ignoring potential relationships among videos. In this work, we revisit AIGC-VQA from an inter-video perspective and formulate it as a reference-aware evaluation problem. Through this formulation, quality assessment is guided not only by intrinsic video characteristics but also by comparisons with related videos, which is more consistent with human perception. To validate its effectiveness, we propose Reference-aware Video Quality Assessment (RefVQA), which utilizes a query-centered reference graph to organize semantically related samples and performs graph-guided difference aggregation from the reference nodes to the query node. Experiments on existing datasets demonstrate that our proposed RefVQA outperforms state-of-the-art methods across multiple quality dimensions, with strong generalization ability validated by cross-dataset evaluation. These results highlight the effectiveness of the proposed reference-based formulation and suggest its potential to advance AIGC-VQA.
\end{abstract}


%

\keywords{AI-generated videos, quality assessment, reference-aware, graph-guided difference aggregation}


\maketitle

\section{Introduction}
\label{sec:intro}

Recent advances in generative models have enabled the rapid proliferation of AI-generated videos (AIGVs) across a wide range of applications, including content creation, media production, and virtual environments \cite{kou2024subjective, wang2025aigv}. As such content becomes increasingly prevalent, the assessment of its visual quality has become a critical problem to ensure reliability and user experience \cite{min2024perceptual}. Unlike traditional video distortions caused by compression or transmission, AIGVs often exhibit more complex artifacts, such as structural inconsistencies, motion instability, and semantic misalignment with the conditioning prompt \cite{lu2024aigc, zhang2024benchmarking}. These characteristics make the automatic quality assessment of such generated videos substantially more challenging, highlighting the need for effective AI-generated content video quality assessment (AIGC-VQA) methods.

\begin{figure}
	\centering
	\vspace{-2pt}
	\includegraphics[width=0.47\textwidth]{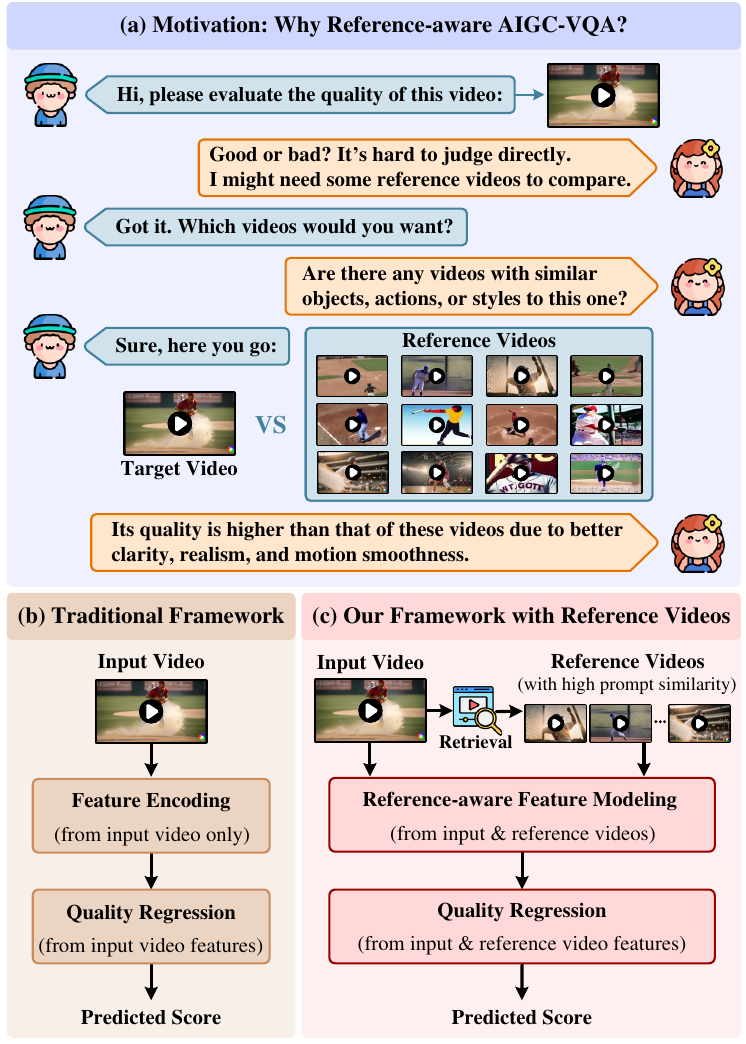}\\
	\vspace{-7pt}
	\caption{Motivation and formulation of our reference-aware AIGC-VQA framework.}
	\vspace{-12pt}
	\label{fig_intro}
\end{figure}

Given the multidimensional nature of quality degradation in AIGVs, existing methods capture features such as motion smoothness and semantic consistency through multi-branch networks \cite{lu2024aigc, qu2024exploring}. To improve the feature decoding capability of the regressor, some methods use large language models (LLMs) for feature decoding and perform a series of cue word optimization \cite{qi2025towards, zhang2025q}. Other studies, based on the statistical distribution and visual patterns of the samples, use techniques such as reinforcement learning \cite{zhang2025vq}, staged learning \cite{lu2024aigc}, and prototype learning \cite{zhang2025lmvq} to improve training and supervision signals. Despite these advances, most existing methods estimate video quality by independently analyzing each sample, making it difficult to reflect the actual subjective quality assessment process.

As illustrated in Figure~\ref{fig_intro}, human perception of video quality is often comparative rather than absolute \cite{zhang2025lmvq}. When evaluating the quality of an input video, observers rarely judge it in isolation. Instead, they implicitly compare it with other videos that share similar objects, actions, or visual styles \cite{lu2024aigc}. Such comparisons provide valuable references that help reveal perceptual differences in clarity, motion smoothness, and semantic consistency.

Motivated by this observation, we reformulate AIGC-VQA as a reference-aware evaluation problem, where the target video is evaluated in the context of semantically similar reference videos. This formulation enables the model to capture relative perceptual differences that are difficult to infer from a single sample. To verify its effectiveness, we propose Reference-aware Video Quality Assessment (RefVQA), a new AIGC-VQA framework that explicitly models reference relationships among generated videos. Since AIGVs are conditioned on textual prompts, these prompts provide a stable, retrievable, and cross-model shared semantic anchor for identifying related videos \cite{cheng2025vpo}. We therefore retrieve candidate references based on prompt similarity and further select valuable samples through similarity ranking. Notably, prompt similarity is used to establish semantic comparability rather than to assume quality equivalence. Compared to visual features, it is less affected by generation quality, providing a more robust basis for reference construction. The retrieved videos are organized into a lightweight reference graph, where nodes correspond to encoded video features extracted by backbone encoders \cite{kou2024subjective} and edges reflect semantic relevance between nodes, serving as a structured reference context.

Although several methods use pairwise or rank losses to model quality comparisons \cite{wang2025aigv, qi2025towards}, they treat comparison as a supervision signal over independently encoded samples. As a result, sample relationships are imposed only at the optimization stage, while the learned features remain largely single-sample based. In contrast, RefVQA embeds reference relationships directly into feature modeling, conditioning the quality representation on semantically related samples and capturing relative quality differences during representation learning rather than enforced only through output-level constraints. Specifically, instead of directly aggregating reference features, RefVQA jointly models the intrinsic features of the target video and its reference-aware difference features under the reference graph topology, thus integrating intrinsic and relative quality cues into a unified representation for quality prediction. In particular, this modeling strategy is seamlessly integrated into both visual and video-prompt alignment branches, allowing the model to capture the two key aspects of AIGC-VQA \cite{kou2024subjective} within a unified framework. To the best of our knowledge, this is the first work to incorporate reference relationships into representation learning for AIGC-VQA through graph-guided difference aggregation.

We conduct extensive experiments on three benchmark datasets, T2VQA-DB \cite{kou2024subjective}, LGVQ \cite{zhang2024benchmarking}, and FETV \cite{liu2023fetv}. The results demonstrate that our method outperforms state-of-the-art approaches across multiple evaluation metrics and quality dimensions. Cross-dataset experiments further verify its generalization capability. In addition, ablation studies and visualization analyses provide clear evidence of the effectiveness of our reference-aware comparative modeling and the rational design of each component. Our main contributions are summarized as follows:

(1) We reformulate AIGC-VQA as a reference-aware evaluation problem, shifting the focus from isolated quality regression to comparative quality reasoning with related samples.

(2) We propose RefVQA, which learns reference-conditioned discrepancy representations in both visual and alignment spaces, enabling the model to capture relative perceptual and alignment quality cues beyond intrinsic video features alone.

(3) We introduce a lightweight graph-guided difference aggregation mechanism to organize retrieved references and encode query-reference differences under a structured topology.

(4) Extensive experiments on existing benchmark datasets demonstrate that RefVQA outperforms state-of-the-art approaches. Furthermore, we conduct ablation studies to analyze the contributions of individual components, and cross-dataset evaluations also show the generalization ability of the proposed method.

\begin{figure*}
	\centering
	\includegraphics[width=0.9\textwidth]{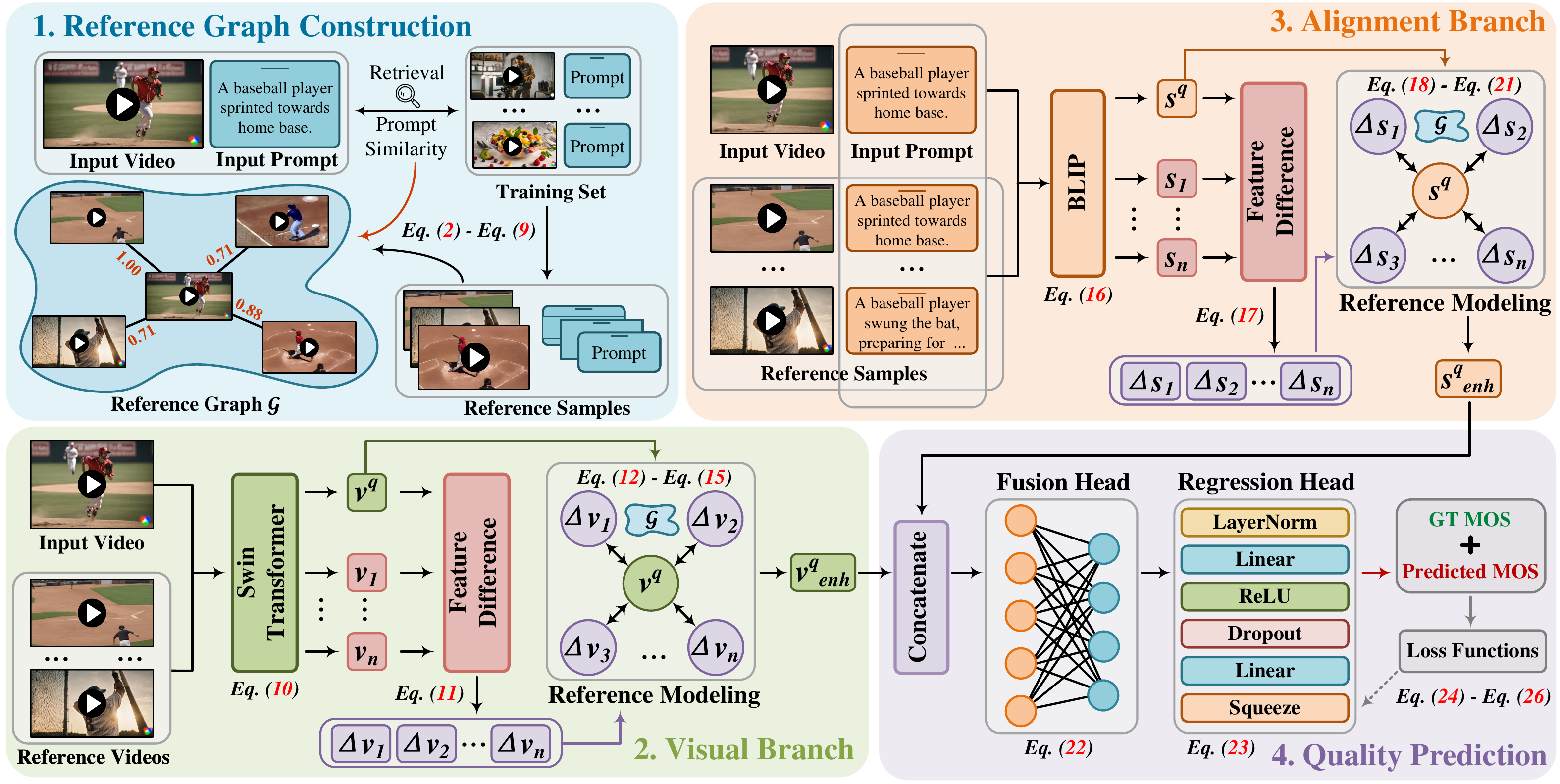}\\
	\vspace{-5pt}
	\caption{Overview of the proposed RefVQA framework. Given an input video and prompt, semantically related reference videos are retrieved to construct a query-centered reference graph. The model then performs graph-guided difference aggregation in both visual and alignment branches, whose outputs are fused for final quality prediction.}
	\vspace{-10pt}
	\label{fig_method}
\end{figure*}

\section{Related Work}
\label{sec:rw}

\subsection{Quality Assessment of AI-Generated Videos}
\label{ssec:qa}
Early studies on VQA for user-generated content primarily focused on perceptual factors such as fidelity, naturalness, and temporal coherence \cite{9633248, wu2023exploring, 9405420}. With the emergence of AIGVs, however, new distortion types have appeared, including object unrealism, motion artifacts, and semantic inconsistencies \cite{qi2025t2veval}. These distortions make the quality assessment of generated videos inherently multi-dimensional \cite{min2024perceptual}, requiring models to capture heterogeneous cues that jointly influence human judgments.

To account for this multi-dimensional nature, many recent AIGC-VQA methods adopt multi-branch architectures that explicitly separate different aspects of perceived quality \cite{lu2024aigc, qi2025t2veval}. A common design employs parallel branches to extract complementary information: a visual branch captures intrinsic appearance and motion characteristics, while an alignment branch evaluates the alignment between generated content and its conditioning prompt \cite{sun2025content, kou2024subjective}. Beyond architectural design, recent work has explored alternative modeling strategies. For example, LLMs have been introduced as replacements for conventional regressors by reformulating AIGC-VQA as a visual question answering problem \cite{wang2025aigv, chen2025vquala}. Other studies improve training strategies and supervision signals through techniques such as synthetic negative sampling \cite{zou2026mds, wu2022fast}, agent-based semantic learning \cite{zhang2025vq}, and instruction-guided optimization \cite{qi2025towards}.


Nevertheless, most existing methods predict the quality of each video independently, relying only on features extracted from the target sample. This overlooks inter-sample comparisons, which reveal differences in visual quality and semantic consistency and provide valuable cues for relative quality assessment.

\subsection{Relational Modeling in Video Analysis}
\label{ssec:rm}

Relational modeling captures dependencies among multiple entities or samples rather than treating them independently \cite{gao2023generalized}. In visual analysis tasks, it has been widely explored to enhance representation learning by exploiting contextual relationships among visual elements \cite{wei2025dreamrelation, xiao2022video}. For example, temporal relational reasoning frameworks model interactions across frames to capture complex motion patterns and event structures in videos \cite{liu2023cross, bai2024event}. Similarly, graph-based architectures represent visual entities as nodes and their interactions as edges, enabling information propagation through structured relationships \cite{zhu2022relational, qiu2025step}.

Graph neural networks (GNNs) have emerged as a widely adopted framework for relational modeling in visual understanding \cite{chen2024survey}. In video analysis, GNN-based approaches model structured dependencies among frames, objects, and semantic components, improving tasks such as action recognition \cite{xing2023boosting}, video understanding \cite{zhou2025videognn}, and visual reasoning \cite{wang2023vqa}. Beyond intra-video relationships, some studies extend relational modeling to inter-sample settings by constructing graphs across videos based on feature similarity or semantic relationships \cite{you2020graph, zeng2021multi}. Information propagation across related samples enables these methods to exploit contextual cues and improve representation robustness \cite{sang2020context}.

Despite these advances, inter-video relationships remain underexplored in AIGC-VQA. In contrast, we construct a reference graph based on semantic similarity between prompts to model cross-video dependencies. Unlike conventional approaches that aggregate features of related samples, we aggregate feature differences between the query and its references, which better aligns with the comparative nature of quality assessment.

\section{Method}
\label{sec:me}

\subsection{Reference-aware Formulation}
\label{sec:for}
Existing AIGC-VQA methods typically predict the quality of each video independently. However, human perception of video quality is inherently comparative. In practice, observers rarely assess a video in isolation, but instead judge it relative to other samples. Such comparisons provide essential cues for perceiving differences in visual sharpness, motion smoothness, and video-prompt semantic consistency, among other factors. Moreover, the use of correlation and ranking-based evaluation metrics \cite{kou2024subjective, qi2025towards} in AIGC-VQA further reflects this comparative nature.

Motivated by this observation, we reformulate AIGC-VQA as a reference-aware evaluation problem. Under this formulation, video quality is assessed not only from intrinsic features of the target video, but also from its perceptual differences with semantically related references. This formulation enables the model to jointly capture absolute and relative quality cues, resulting in predictions that better align with human quality perception.

Formally, let $S = (V, T)$ denote a video-prompt pair. Conventional AIGC-VQA predicts quality of each sample independently, while our formulation incorporates reference-aware differences:
\begin{equation}
	y = f\big(S,\; \Delta(S, \mathcal{R}(S))\big),
\end{equation}
where $y$ denotes the predicted quality score, $f(\cdot)$ is a quality prediction function, $\mathcal{R}(\cdot)$ denotes reference retrieval, and $\Delta(\cdot)$ captures relative differences.

\subsection{Overview of RefVQA}
\label{sec:ov}

To realize the proposed reference-aware formulation, we introduce RefVQA, a reference-aware AIGC-VQA framework. As illustrated in Figure~\ref{fig_method}, RefVQA consists of four modules. First, the \textbf{Reference Graph Construction} module retrieves semantically related videos based on prompt similarity and organizes them into a query-centered reference graph, where nodes correspond to video instances and edges encode the semantic similarity between their prompts. This graph defines the relational structure among videos and provides the topology for subsequent feature modeling. Second, the \textbf{Visual Branch} extracts spatiotemporal representations from videos to characterize visual appearance and motion patterns. Third, the \textbf{Alignment Branch} employs BLIP \cite{li2022blip} to model the semantic consistency between generated videos and their conditioning prompts. In both branches, the shared graph topology is used for graph-guided difference aggregation to generate enhanced quality features. Finally, the \textbf{Quality Prediction} module integrates these features to produce a unified quality representation, which is then used to predict the final quality score.

\subsection{Reference Graph Construction}
\label{sec:rgc}

Given a query sample $(V^{q}, T^{q})$, where $V^{q}$ denotes the generated video and $T^{q}$ is the corresponding prompt, we retrieve its semantically related samples to construct a query-centered reference graph.

Specifically, let the training set be denoted as:
\begin{equation}
	\mathcal{D} = \{(V_n, T_n)\}_{n=1}^{N},
\end{equation}
where $N$ denotes the number of training samples.

To measure semantic similarity between prompts, we encode both the query prompt and the prompts in the training set:
\begin{equation}
	\mathbf{e}^{q} = f_{\text{bert}}(T^{q}), \quad
	\mathbf{e}_n = f_{\text{bert}}(T_n), \quad n=1,\ldots,N,
\end{equation}
where $f_{\text{bert}}(\cdot)$ denotes the Sentence-BERT \cite{reimers2019sentence} encoder. The similarity between the query prompt and each training prompt is then computed using cosine similarity:
\begin{equation}
	s_n = \mathrm{sim}(\mathbf{e}^{q}, \mathbf{e}_n) =
	\frac{(\mathbf{e}^{q})^{\top} \mathbf{e}_n}{\|\mathbf{e}^{q}\|_2 \, \|\mathbf{e}_n\|_2},
	\quad n = 1,\ldots,N .
	\label{eq:similarity}
\end{equation}

Based on the similarity scores, we select reference indices whose similarity exceeds a predefined threshold $\tau$:
\begin{equation}
	\mathcal{N}^{q} = \{\, n \mid s_n > \tau,\; V_n \neq V^{q} \,\}.
	\label{eq:neighbor}
\end{equation}

The corresponding reference set is therefore defined as:
\begin{equation}
	\mathcal{R}^{q} = \{ (V_n, T_n) \mid n \in \mathcal{N}^{q} \}.
\end{equation}

Using the query sample and its retrieved references, we construct a query-centered reference graph:
\begin{equation}
	\mathcal{G}^{q} = (\mathcal{V}^{q}, \mathcal{E}^{q}),
\end{equation}
where $\mathcal{V}^{q}$ and $\mathcal{E}^{q}$ denote the node set and edge set, respectively. The node set is defined as:
\begin{equation}
	\mathcal{V}^{q} = \{ V^{q} \} \cup \{ V_n \mid n \in \mathcal{N}^{q} \}.
\end{equation}

Each node corresponds to a video sample, including the query video and its retrieved reference videos. The graph only contains edges between the query node and its reference nodes \cite{pan2020star}. Accordingly, the edge set is defined as:
\begin{equation}
	\mathcal{E}^{q} = \{\, (V^{q}, V_n) \mid n \in \mathcal{N}^{q} \,\}.
\end{equation}

Each edge $(V^{q}, V_n)$ is assigned the weight $s_n$, which is the prompt similarity defined in Eq.~(\ref{eq:similarity}). In this way, the constructed graph captures the semantic relevance between the query sample and its retrieved references, and provides a topology for the subsequent graph-guided difference aggregation.

\subsection{Visual Branch}
\label{sec:vb}

The visual branch is designed to capture spatio-temporal perceptual cues, enabling the assessment of both visual appearance and motion quality in generated videos \cite{kou2024subjective}.

Given an input video $V^{q}$ and its retrieved reference videos $\{V_n \mid n \in \mathcal{N}^{q}\}$, we extract their visual features using a Swin Transformer \cite{Liu_2022_CVPR} encoder $f_{\text{swin}}(\cdot)$ followed by a 3D adaptive average pooling layer $f^{\text{v}}_{\text{pool}}(\cdot)$:
\begin{equation}
	\mathbf{v}^{q} = f^{\text{v}}_{\text{pool}}(f_{\text{swin}}(V^{q})), \quad 
	\mathbf{v}_n = f^{\text{v}}_{\text{pool}}(f_{\text{swin}}(V_n)), \quad n \in \mathcal{N}^{q},
\end{equation}
where $\mathbf{v}^{q}, \mathbf{v}_n \in \mathbb{R}^{D_v}$ denote the resulting visual features. To explicitly characterize the discrepancy between the input and reference videos, we construct reference-aware difference features:
\begin{equation}
	\Delta \mathbf{v}_n = \mathbf{v}^q - \mathbf{v}_n, \quad n \in \mathcal{N}^{q},
\end{equation}
which encourages the model to focus on visual differences instead of shared content. Rather than performing full graph propagation, we adopt a query-centered aggregation scheme that only collects messages from reference nodes to the query node \cite{pan2020star}. The aggregated reference feature $\mathbf{d}^{q}\in \mathbb{R}^{D_v}$ is computed as:
\begin{equation}
	\mathbf{d}^{q} = \mathrm{GeLU} \left(
	\sum_{n \in \mathcal{N}^{q}} s_n \, \mathbf{W}_\text{v} \Delta \mathbf{v}_n
	\right),
\end{equation}
where $\mathbf{W}_\text{v} \in \mathbb{R}^{D_v \times D_v}$ is a learnable projection matrix and $\mathrm{GeLU}(\cdot)$ denotes the Gaussian error linear unit.

We further refine both $\mathbf{v}^{q}$ and $\mathbf{d}^{q}$ using lightweight adapters to enhance their compatibility for subsequent fusion:
\begin{equation}
	\tilde{\mathbf{v}}^{q} = f^{\text{v}}_{\text{adapt}}(\mathbf{v}^{q}), \qquad 
	\tilde{\mathbf{d}}^{q} = f^{\text{d}}_{\text{adapt}}(\mathbf{d}^{q}),
\end{equation}
where $f^{\text{v}}_{\text{adapt}}(\cdot)$ and $f^{\text{d}}_{\text{adapt}}(\cdot)$ are trainable adapters implemented as Multi-Layer Perceptrons (MLPs) with residual connections.

To adaptively regulate the contribution of reference information, we introduce a gating mechanism:
\begin{equation}
	\alpha_{\text{v}} = \sigma \left( \mathbf{w}_\text{v}^{\top} [\tilde{\mathbf{v}}^{q} \, \Vert \, \tilde{\mathbf{d}}^{q}] \right),
\end{equation}
where $\sigma(\cdot)$ denotes the sigmoid function, $\mathbf{w}_\text{v} \in \mathbb{R}^{2D_v}$ is a learnable projection vector, and $[\cdot \Vert \cdot]$ denotes feature concatenation.

Finally, the reference-enhanced visual representation $\mathbf{v}^{q}_{\text{enh}}\in \mathbb{R}^{D_v}$ is obtained by:
\begin{equation}
	\mathbf{v}^{q}_{\text{enh}} = \mathrm{LN} \left( f_{\text{mlp}}^\text{v} \big( [\tilde{\mathbf{v}}^{q} \, \Vert \, \alpha_\text{v} \tilde{\mathbf{d}}^{q}] \big) \right),
\end{equation}
where $f_{\text{mlp}}^\text{v}(\cdot)$ is an MLP and $\mathrm{LN}(\cdot)$ denotes Layer Normalization.

\subsection{Alignment Branch}
\label{sec:sb}

The alignment branch is designed to capture cross-modal alignment between generated videos and their corresponding prompts, facilitating the assessment of video-prompt alignment.

Given an input sample $(V^{q}, T^{q})$ and its reference set $\mathcal{R}^{q}$, we encode each sample as:
\begin{equation}
	\begin{aligned}
		\mathbf{s}^{q} &= f^{\text{s}}_{\text{pool}}(f_{\text{blip}}(V^{q}, T^{q})), \\
		\mathbf{s}_n &= f^{\text{s}}_{\text{pool}}(f_{\text{blip}}(V_n, T_n)), (V_n, T_n) \in \mathcal{R}^{q},
	\end{aligned}
\end{equation}
where $\mathbf{s}^{q}, \mathbf{s}_n \in \mathbb{R}^{D_s}$ denote the resulting alignment features, $f_{\text{blip}}(\cdot)$ denotes the BLIP encoder, and $f^{\text{s}}_{\text{pool}}(\cdot)$ is a 2D adaptive average pooling layer.

Similar to the visual branch, we construct alignment difference features between the input and reference samples:
\begin{equation}
	\Delta \mathbf{s}_n = \mathbf{s}^{q} - \mathbf{s}_n, \quad n \in \mathcal{N}^{q}.
\end{equation}

Following the graph topology defined in Sec.~\ref{sec:rgc}, the aggregated reference feature $\mathbf{g}^{q}\in \mathbb{R}^{D_s}$ is computed as:
\begin{equation}
	\mathbf{g}^{q} =
	\mathrm{GeLU} \left(
	\sum_{n \in \mathcal{N}^{q}} s_n \mathbf{W}_\text{s} \Delta \mathbf{s}_n
	\right),
\end{equation}
where $\mathbf{W}_\text{s} \in \mathbb{R}^{D_s \times D_s}$ is a learnable projection matrix.

Finally, the reference-enhanced alignment representation of the input sample is obtained by adaptively combining its intrinsic alignment feature with the aggregated reference feature:
\begin{equation}
	\tilde{\mathbf{s}}^{q} = f^{\text{s}}_{\text{adapt}}(\mathbf{s}^{q}), \qquad 
	\tilde{\mathbf{g}}^{q} =f^{\text{g}}_{\text{adapt}}({\mathbf{g}^{q}}),
\end{equation}
\begin{equation}
	\alpha_\text{s} =
	\sigma \left(
	\mathbf{w}_\text{s}^{\top} [\tilde{\mathbf{s}}^{q} \Vert \tilde{\mathbf{g}}^{q}]
	\right),
\end{equation}
\begin{equation}
	\mathbf{s}^{q}_{\text{enh}} = \mathrm{LN} \left( f_{\text{mlp}}^\text{s} \big( [\tilde{\mathbf{s}}^{q} \, \Vert \, \alpha_\text{s} \tilde{\mathbf{g}}^{q}] \big) \right),
\end{equation}
where $f^{\text{s}}_{\text{adapt}}(\cdot)$ and $f^{\text{g}}_{\text{adapt}}(\cdot)$ are adapters,  $\mathbf{w}_\text{s} \in \mathbb{R}^{2D_s}$ is learnable projection vector, and $ f_{\text{mlp}}^\text{s}$ denotes an MLP.

\subsection{Quality Prediction}

In this module, we adopt a lightweight prediction structure for quality estimation to minimize its impact and better isolate the effect of reference-aware comparative modeling across branches.

We first fuse the visual and alignment features through a fusion head to obtain the unified quality representation $\mathbf{h}^{q} \in \mathbb{R}^{D_h}$:
\begin{equation}
	\mathbf{h}^{q} = f_{\text{c}} \big( [\mathbf{v}^{q}_{\text{enh}} \, \Vert \, \mathbf{s}^{q}_{\text{enh}}] \big),
\end{equation}
where $f_{\text{c}}(\cdot)$ is implemented as an MLP with ReLU activations and dropout layers.

The final quality score $y^q$ is then predicted by a regression head $f_{\text{reg}}(\cdot)$ with a softplus activation to ensure non-negativity:
\begin{equation}
	y^q = \log\big(1 + \exp(f_{\text{reg}}(\mathbf{h}^{q}))\big).
\end{equation}

\section{Experiments}
\label{sec:exp}

\subsection{Datasets}

We conduct experiments on three widely adopted benchmarks for AIGC-VQA, namely T2VQA-DB \cite{kou2024subjective}, LGVQ \cite{zhang2024benchmarking}, and FETV \cite{liu2023fetv}. T2VQA-DB contains 10,000 videos generated by 9 text-to-video (T2V) models, each annotated with a mean opinion score (MOS) collected from 27 human raters. LGVQ consists of 2,808 videos synthesized by 6 T2V models based on 468 prompts, while FETV includes 2,476 videos generated from 619 prompts using 4 models. In contrast to T2VQA-DB, both LGVQ and FETV provide multi-dimensional annotations, covering spatial quality, temporal quality, and text-video alignment. Following the official protocols, each dataset is randomly divided into 80\% training and 20\% testing subsets. To mitigate the variance introduced by a single split, we repeat the experiment five times with different random seeds and report the averaged results \cite{kou2024subjective}.

\subsection{Metrics}

To comprehensively assess prediction performance, we adopt four standard metrics: Pearson linear correlation coefficient (PLCC), Spearman rank-order correlation coefficient (SRCC), Kendall’s rank-order correlation coefficient (KRCC), and root mean square error (RMSE), following prior studies \cite{radford2021learning, wu2022disentangling}. PLCC evaluates the linear agreement between predicted scores and ground-truth (GT) MOS, whereas SRCC and KRCC measure the consistency in relative ranking. RMSE reflects the magnitude of prediction errors. Superior models are expected to achieve higher correlation coefficients (PLCC, SRCC, KRCC) and lower RMSE values.

\subsection{Implementation Details}

Following T2VQA \cite{kou2024subjective}, the Swin Transformer is pretrained on Kinetics-400 \cite{kay2017kinetics}, while BLIP and Sentence-BERT load their official pretrained weights. Each video is uniformly sampled into 8 frames and resized to $224 \times 224$. For reference retrieval, we construct a reference pool using the training set, which is shared during both training and inference, without using any test samples. The retrieval threshold $\tau$ is set to 0.7. Inspired by previous works \cite{qi2025t2veval, lu2024aigc}, we adopt a decoupled optimization strategy to balance performance and computational efficiency. Optimization is performed using AdamW with a weight decay of 0.05, along with a linear warm-up schedule followed by cosine annealing.

In the first training stage, the entire framework is jointly optimized to adapt the pretrained backbones to the AIGC-VQA task. We adopt a layer-wise learning rate scheme, setting $1\times10^{-5}$ for the backbones and $5\times10^{-6}$ for the remaining modules. This stage runs for 10 epochs. In the second stage, the backbones are frozen and their output features are pre-extracted and cached offline, enabling efficient reference retrieval without repeated backbone inference. The remaining modules are further optimized with a learning rate of $1\times10^{-5}$ for 20 epochs. All experiments are conducted on two NVIDIA RTX 4090 GPUs with a batch size of $B=8$ per GPU.

To better align predictions with human subjective perception, we optimize the model using a hybrid objective that combines PLCC loss and rank loss \cite{qi2025towards}. Specifically, the PLCC loss measures linear consistency between predictions and MOS values, while the rank loss encourages correct pairwise ordering among samples. The overall objective is defined as:
\begin{equation}
	L_{\mathrm{plcc}} = \frac{1}{2} \left(
	1 -
	\frac{\sum_{i=1}^{m} (y_i - \bar{y})(s_i - \bar{s})}
	{\sqrt{\sum_{i=1}^{m} (y_i - \bar{y})^2}
		\sqrt{\sum_{i=1}^{m} (s_i - \bar{s})^2}}
	\right),
	\label{eq:plcc_loss}
\end{equation}
\begin{equation}
	L_{\mathrm{rank}} =
	\frac{1}{m^2}
	\sum_{i=1}^{m} \sum_{j=1}^{m}
	\max \left(
	0,\,
	|s_i - s_j| - e(s_i, s_j)\cdot (y_i - y_j)
	\right),
	\label{eq:rank_loss}
\end{equation}
\begin{equation}
	L = L_{\mathrm{plcc}} + \gamma \cdot L_{\mathrm{rank}},
	\label{eq:total_loss}
\end{equation}
\noindent where $m$ denotes the number of videos in a mini-batch, $y_i$ and $s_i$ represent the predicted score and the GT MOS of the $i$-th video, respectively, and $\bar{y}$ and $\bar{s}$ denote their corresponding mean values within the mini-batch. $e(s_i, s_j)$ denotes the relative ordering \cite{qi2025towards} between $s_i$ and $s_j$. Meanwhile, the hyper-parameter $\gamma$ balances the contributions of the two loss terms and is set to 0.3 \cite{kou2024subjective}.

\subsection{Performance Comparison}

\paragraph{\textbf{In-domain Evaluation.}} Tables~\ref{tab:T2VQA} and \ref{tab:LGVQ} report the experimental results on T2VQA-DB and LGVQ, respectively. RefVQA consistently outperforms existing methods on both datasets. On T2VQA-DB, it achieves the best SRCC (0.826), PLCC (0.835), KRCC (0.642), and RMSE (8.429) values. On LGVQ, RefVQA achieves state-of-the-art results on spatial, temporal, and alignment dimensions, demonstrating more balanced performance across multiple dimensions compared with methods that focus on improving a single aspect. Notably, the gain is particularly significant for alignment, indicating that retrieving semantically relevant references provides stronger cues for assessing video-prompt alignment. These results demonstrate that the proposed method achieves consistently strong performance across multiple datasets and evaluation dimensions, highlighting its robustness and general applicability for AIGC-VQA.

\begin{table}[htbp]
	\caption{Performance comparison on T2VQA-DB.}
	\vspace{-5pt}
	\label{tab:T2VQA}
	\centering
	\resizebox{0.47\textwidth}{!}{
		\begin{tabular}{cccccc}
			\toprule
			\textbf{Method} & \textbf{Year} & \textbf{SRCC $\uparrow$} & \textbf{PLCC $\uparrow$} & \textbf{KRCC $\uparrow$} & \textbf{RMSE $\downarrow$}\\
			\midrule
			CLIPSim \cite{radford2021learning} & ICML 21 & 0.105 & 0.128 & 0.070 & 21.683\\
			SimpleVQA \cite{sun2022deep} & ACM MM 22 & 0.628 & 0.634 & 0.447 & 11.163\\
			DOVER \cite{wu2022disentangling} & arXiv 22 & 0.761 & 0.769 & 0.570 & 9.807\\
			FAST-VQA \cite{wu2022fast} & ECCV 22 & 0.717 & 0.730 & 0.530 & 10.595\\
			BLIP \cite{li2022blip} & ICML 22 & 0.166 & 0.186 & 0.111 & 18.373\\
			BVQA \cite{li2022blindly} & TCSVT 22 & 0.739 & 0.749 & 0.549 & 15.645\\
			KSVQE \cite{madhusudana2022image} & TIP 22 & 0.771 & 0.784 & 0.594 & 11.053\\
			ImageReward \cite{xu2023imagereward} & NeurIPS 23 & 0.188 & 0.212 & 0.127 & 18.243\\
			UMTScore \cite{liu2023fetv} & NeurIPS 23 & 0.067 & 0.072 & 0.045 & 22.559\\
			ViCLIP \cite{wang2023internvid} & ICLR 24 & 0.116 & 0.145 & 0.078 & 21.655\\
			T2VQA \cite{kou2024subjective} & ACM MM 24 & 0.797 & 0.807 & 0.606 & 9.022\\
			Zhang et al. \cite{yue2024advancing} & arXiv 24 & 0.742 & 0.757 & 0.555 & -\\
			VE-Bench QA \cite{sun2025ve} & AAAI 2025 & 0.818 & 0.823 & 0.637 & -\\
			T2VEval \cite{qi2025t2veval} & arXiv 25 & 0.805 & 0.818 & 0.616 & 8.613\\
			CRAVE \cite{sun2025content} & arXiv 25 & 0.812 & 0.821 & 0.634 & -\\
			AIGVEval \cite{qi2025towards} & CVPRW 25 & 0.763 & 0.749 & 0.561 & 10.034\\
			MSA-VQA \cite{li2025multilevel} & ICASSP 25 & 0.810 & 0.825 & - & -\\
			LMVQ \cite{zhang2025lmvq} & TCSVT 25 & 0.772 & 0.784 & 0.597 & - \\
			\rowcolor[HTML]{E6E6FA} 
			RefVQA (Ours) & - & \textbf{0.826} & \textbf{0.835} & \textbf{0.642} & \textbf{8.429}\\
			\bottomrule
		\end{tabular}
	}
	\vspace{-8pt}
\end{table}

\paragraph{\textbf{Cross-dataset Generalization.}} To further evaluate the generalization capability of the proposed model, following \cite{zhang2024benchmarking}, we conduct cross-dataset experiments by training on LGVQ and testing on FETV. As shown in Table~\ref{tab:FETV}, RefVQA consistently outperforms all competing methods across spatial, temporal, and alignment dimensions, with particularly significant improvements on temporal and alignment quality. This strong generalization performance can be attributed to the reference-aware design, which performs relative quality modeling based on semantically similar samples rather than relying on dataset-specific feature distributions. Consequently, the proposed method is less sensitive to domain shifts and exhibits strong generalization to unseen data.

\begin{table}[t]
	\caption{Performance comparison on LGVQ.}
	\vspace{-7pt}
	\label{tab:LGVQ}
	\centering
	\resizebox{0.47\textwidth}{!}{
		\begin{tabular}{cccccc}
			\toprule
			\textbf{Type} & \textbf{Method} & \textbf{Year} & \textbf{SRCC $\uparrow$} & \textbf{PLCC $\uparrow$} & \textbf{KRCC $\uparrow$}\\
			\midrule
			\multirow{9}{*}{Spatial} & MUSIQ \cite{ke2021musiq} & ICCV 21 & 0.669 & 0.682 & 0.491\\
			& UNIQUE \cite{zhang2021uncertainty} & TIP 21 & 0.716 & 0.525 & 0.768\\
			& StairIQA \cite{sun2023blind} & JSTSP 23 & 0.701 & 0.737 & 0.521\\
			& CLIP-IQA \cite{wang2023exploring} & AAAI 23 & 0.684 & 0.709 & 0.502\\
			& LIQE \cite{zhang2023blind} & CVPR 23 & 0.721 & 0.752 & 0.538\\
			& UGVQ \cite{zhang2024benchmarking} & TOMM 25 & {0.759} & {0.795} & {0.567}\\
			\rowcolor[HTML]{E6E6FA} 
			& RefVQA (Ours) & - & \textbf{0.764} & \textbf{0.803} & \textbf{0.576}\\
			\midrule
			\multirow{7}{*}{Temporal} & SimpleVQA \cite{sun2022deep} & ACM MM 22 & 0.857 & 0.867 & 0.659\\
			& FastVQA \cite{wu2022fast} & ECCV 22 & 0.849 & 0.843 & 0.647\\
			& DOVER \cite{wu2023exploring} & ICCV 23 & 0.867 & 0.878 & 0.672\\
			& UGVQ \cite{zhang2024benchmarking} & TOMM 25 & {0.893} & {0.907} & {0.703}\\
			\rowcolor[HTML]{E6E6FA} 
			& RefVQA (Ours) & - & \textbf{0.906} & \textbf{0.924} & \textbf{0.723}\\
			\midrule
			\multirow{9}{*}{Alignment} & CLIPScore \cite{hessel2021clipscore} & EMNLP 21 & 0.446 & 0.453 & 0.301\\
			& BLIPScore \cite{li2022blip} & ICML 22 & 0.455 & 0.464 & 0.319\\
			& ImageReward \cite{xu2023imagereward} & NeurIPS 23 & 0.498 & 0.499 & 0.344\\
			& PickScore \cite{kirstain2023pick} & NeurIPS 23 & 0.501 & 0.515 & 0.353\\
			& HPSv2 \cite{wu2023human} & arXiv 23 & 0.504 & 0.511 & 0.357\\
			& UGVQ \cite{zhang2024benchmarking} & TOMM 25 & {0.551} & {0.555} & {0.394}\\
			\rowcolor[HTML]{E6E6FA}
			& RefVQA (Ours) & - & \textbf{0.731} & \textbf{0.738} & \textbf{0.544}\\
			\bottomrule
		\end{tabular}
	}
	\vspace{-12pt}
\end{table}

\begin{table}[htbp]
	\caption{Cross-dataset performance comparison (trained on LGVQ and tested on FETV).}
	\vspace{-5pt}
	\label{tab:FETV}
	\centering
	\resizebox{0.47\textwidth}{!}{
		\begin{tabular}{cccccc}
			\toprule
			\textbf{Type} & \textbf{Method} & \textbf{Year} & \textbf{SRCC $\uparrow$} & \textbf{PLCC $\uparrow$} & \textbf{KRCC $\uparrow$}\\
			\midrule
			\multirow{7}{*}{Spatial} & MUSIQ \cite{ke2021musiq} & ICCV 21 & 0.426 &  0.472 & 0.312\\
			& UNIQUE \cite{zhang2021uncertainty} & TIP 21 & 0.389 & 0.383 & 0.272\\
			& StairIQA \cite{sun2023blind} & JSTSP 23 & 0.501 & 0.539 & 0.360\\
			& CLIP-IQA \cite{wang2023exploring} & AAAI 23 & 0.503 & 0.543 & 0.361\\
			& LIQE \cite{zhang2023blind} & CVPR 23 & 0.478 & 0.519 & 0.341\\
			& UGVQ \cite{zhang2024benchmarking} & TOMM 25 & {0.553} & {0.555} & {0.406}\\
			\rowcolor[HTML]{E6E6FA} 
			& RefVQA (Ours) & - & \textbf{0.570} & \textbf{0.579} & \textbf{0.421}\\
			\midrule
			\multirow{5}{*}{Temporal} & SimpleVQA \cite{sun2022deep} & ACM MM 22 & 0.501 & 0.511 & 0.366\\
			& FastVQA \cite{wu2022fast} & ECCV 22 & 0.482 & 0.494 & 0.324\\
			& DOVER \cite{wu2023exploring} & ICCV 23 & 0.494 & 0.483 & 0.349\\
			& UGVQ \cite{zhang2024benchmarking} & TOMM 25 & {0.512} & {0.535} & {0.368}\\
			\rowcolor[HTML]{E6E6FA} 
			& RefVQA (Ours) & - & \textbf{0.719} & \textbf{0.715} & \textbf{0.564}\\
			\midrule
			\multirow{7}{*}{Alignment} & CLIPScore \cite{hessel2021clipscore} & EMNLP 21 &  0.234 & 0.252 & 0.167\\
			& BLIPScore \cite{li2022blip} & ICML 22 & 0.216 & 0.233 & 0.144\\
			& ImageReward \cite{xu2023imagereward} & NeurIPS 23 & 0.261 & 0.283 & 0.183\\
			& PickScore \cite{kirstain2023pick} & NeurIPS 23 & 0.259 & 0.284 & 0.178\\
			& HPSv2 \cite{wu2023human} & arXiv 23 &  0.263 & 0.285 & 0.189\\
			& UGVQ \cite{zhang2024benchmarking} & TOMM 25 & {0.278} & {0.292} & {0.196}\\
			\rowcolor[HTML]{E6E6FA} 
			& RefVQA (Ours) & - & \textbf{0.432} & \textbf{0.453} & \textbf{0.310}\\
			\bottomrule
		\end{tabular}
	}
	\vspace{-8pt}
\end{table}

\subsection{Ablation Studies}

\begin{table*}[t]
	\centering
	\setlength{\tabcolsep}{4pt}
	\captionsetup[subtable]{labelformat=parens, justification=centering}
	\vspace{-3pt}
	
	\begin{subtable}[t]{0.32\textwidth}
		\centering
		\parbox[t][1.6em][t]{\linewidth}{\centering\caption{Reference-aware comparative modeling.}}
		{\scriptsize
			\begin{tabular}{cccccc}
				\toprule
				\multicolumn{2}{c}{\textbf{Reference Usage}} & \multicolumn{4}{c}{\textbf{T2VQA-DB}}\\
				\cmidrule(lr){1-2} \cmidrule(lr){3-6}
				Visual & Alignment & SRCC $\uparrow$ & PLCC $\uparrow$ & KRCC $\uparrow$ & RMSE $\downarrow$\\
				\cmidrule{1-6}
				& & 0.805 & 0.807 & 0.616 & 8.993\\
				\checkmark & & \underline{0.819} & \underline{0.826} & \underline{0.632} & \underline{8.550}\\
				& \checkmark & 0.814 & 0.818 & 0.624 & 8.736\\
				\checkmark & \checkmark & \textbf{0.826} & \textbf{0.835} & \textbf{0.642} & \textbf{8.429}\\
				\bottomrule
			\end{tabular}
		}
		\label{tab:r1}
	\end{subtable}
	\hfill
	\begin{subtable}[t]{0.32\textwidth}
		\centering
		\parbox[t][1.6em][t]{\linewidth}{\centering\caption{Reference retrieval strategies.}}
		{\scriptsize
			\begin{tabular}{ccccc}
				\toprule
				\multirow{2}{*}{\textbf{Retrieval Strategy}} & \multicolumn{4}{c}{\textbf{T2VQA-DB}}\\
				\cmidrule(lr){2-5}
				& SRCC $\uparrow$ & PLCC $\uparrow$ & KRCC $\uparrow$ & RMSE $\downarrow$\\
				\cmidrule{1-5}
				Random & 0.809 & 0.816 & 0.618 & 8.783 \\
				Batch & 0.813 & 0.818 & 0.623 & 8.744 \\
				Feature Similarity & \underline{0.819} & \underline{0.823} & \underline{0.629} & \underline{8.613} \\
				Prompt Similarity & \textbf{0.826} & \textbf{0.835} & \textbf{0.642} & \textbf{8.429}\\
				\bottomrule
			\end{tabular}
		}
		\label{tab:r2}
	\end{subtable}
	\hfill
	\begin{subtable}[t]{0.32\textwidth}
		\centering
		\parbox[t][1.6em][t]{\linewidth}{\centering\caption{Reference feature modeling strategies.}}
		
		{\scriptsize
			\begin{tabular}{ccccc}
				\toprule
				\multirow{2}{*}{\textbf{Modeling Strategy}} & \multicolumn{4}{c}{\textbf{T2VQA-DB}}\\
				\cmidrule(lr){2-5}
				& SRCC $\uparrow$ & PLCC $\uparrow$ & KRCC $\uparrow$ & RMSE $\downarrow$\\
				\cmidrule{1-5}
				Self + Avg & 0.812 & 0.815 & 0.621 & 8.816 \\
				Diff + Avg & \underline{0.821} & \underline{0.826} & \underline{0.633} & \underline{8.458} \\
				Self + Graph & 0.815 & 0.818 & 0.625 & 8.753 \\
				Diff + Graph & \textbf{0.826} & \textbf{0.835} & \textbf{0.642} & \textbf{8.429}\\
				\bottomrule
			\end{tabular}
		}
		\label{tab:r3}
	\end{subtable}
	
	\vspace{0.5em}
	
	\begin{subtable}[t]{0.45\textwidth}
		\centering
		\parbox[t][1.6em][t]{\linewidth}{\centering\caption{Branch configuration.}}
		
		{\scriptsize
			\begin{tabular}{cccccc}
				\toprule
				\multicolumn{2}{c}{\textbf{Branch}} & \multicolumn{4}{c}{\textbf{T2VQA-DB}}\\
				\cmidrule(lr){1-2} \cmidrule(lr){3-6}
				Visual & Alignment & SRCC $\uparrow$ & PLCC $\uparrow$ & KRCC $\uparrow$ & RMSE $\downarrow$\\
				\cmidrule{1-6}
				\checkmark & & 0.751 & 0.772 & 0.561 & 9.927\\
				& \checkmark & \underline{0.765} & \underline{0.780} & \underline{0.575} & \underline{9.697}\\
				\checkmark & \checkmark & \textbf{0.826} & \textbf{0.835} & \textbf{0.642} & \textbf{8.429}\\
				\bottomrule
			\end{tabular}
		}
		\label{tab:r4}
	\end{subtable}
	\hfill
	\begin{subtable}[t]{0.54\textwidth}
		\centering
		\parbox[t][1.6em][t]{\linewidth}{\centering\caption{Training and inference strategies.}}
		
		{\scriptsize
			\begin{tabular}{ccccccc}
				\toprule
				\multirow{2}{*}{\textbf{Strategy}} & \multicolumn{6}{c}{\textbf{T2VQA-DB}}\\
				\cmidrule(lr){2-7}
				& SRCC $\uparrow$ & PLCC $\uparrow$ & KRCC $\uparrow$ & RMSE $\downarrow$ & Training Time $\downarrow$ & Inference Time $\downarrow$\\
				\cmidrule{1-7}
				Frozen Backbone & 0.781 & 0.758 & 0.590 & 9.507 & \textbf{9.908} & \textbf{3.366} \\
				Joint Training & \textbf{0.827} & \underline{0.831} & \underline{0.634} & \textbf{8.425} & 95.319 & \underline{8.503} \\
				Ours & \underline{0.826} & \textbf{0.835} & \textbf{0.642} & \underline{8.429} & \underline{38.378} & \textbf{3.336} \\
				\bottomrule
			\end{tabular}
		}
		\label{tab:r5}
	\end{subtable}
	\vspace{0.5em}
	\caption{Ablation studies on key components of the proposed method.}
	\vspace{-10pt}
	\label{tab:ablation_all}
\end{table*}

\begin{figure*}
	\vspace{-5pt}
	\centering	\includegraphics[width=0.9\textwidth]{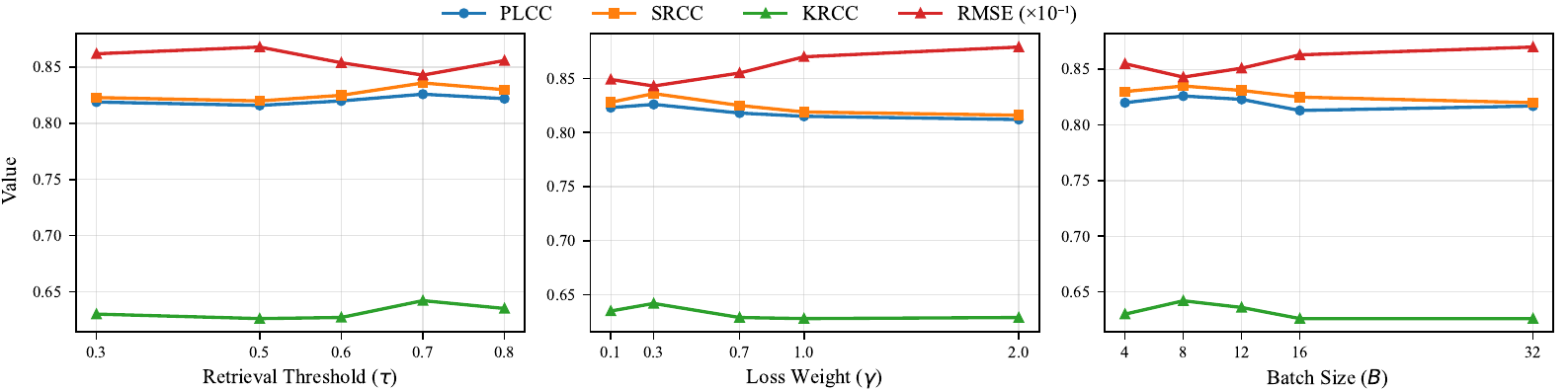}\\
	\vspace{-5pt}
	\caption{Ablation studies on hyperparameters.}
	\vspace{-8pt}
	\label{fig_parameter}
\end{figure*}

We conduct ablation studies to analyze six key aspects of the proposed model, including the use of reference-aware comparative modeling, reference retrieval strategies, reference feature modeling strategies, branch configuration, training and inference strategies, and hyperparameter settings.

\paragraph{\textbf{Reference-aware Comparative Modeling.}} As shown in Table~\ref{tab:ablation_all} (a), introducing reference-aware comparative modeling consistently improves model performance. Applying it to the visual branch yields larger gains than the alignment branch, indicating that reference information is more effective for capturing visual quality differences. Importantly, jointly enabling both branches achieves the best results, demonstrating their complementary roles in AIGC-VQA. This observation is consistent with our reference-aware formulation, where video quality is assessed through comparisons with related samples rather than in isolation.

\paragraph{\textbf{Reference Retrieval Strategy.}} We compare different reference retrieval strategies, including random sampling from the training set (\textit{Random}), in-batch mutual referencing (\textit{Batch}), feature similarity based on backbone embeddings (\textit{Feature Similarity}), and our prompt-based similarity (\textit{Prompt Similarity}). As shown in Table~\ref{tab:ablation_all} (b), retrieval quality plays an important role in performance. While \textit{Random} and \textit{Batch} strategies provide limited gains, similarity-based approaches are more effective. In particular, prompt-based retrieval achieves the best results, outperforming feature similarity, demonstrating that high-level semantic alignment is more beneficial than low-level feature matching for reference selection.

\paragraph{\textbf{Reference feature modeling strategy.}} We further study different reference feature modeling strategies, including using features of reference videos directly (\textit{Self}) or difference features w.r.t. the given video (\textit{Diff}), together with two aggregation schemes over multiple references, i.e., average pooling (\textit{Avg}) and our graph-based aggregation (\textit{Graph}). As shown in Table~\ref{tab:ablation_all} (c), modeling difference features consistently outperforms using raw reference features, indicating the importance of explicit discrepancy modeling. Moreover, graph-based aggregation achieves better results than simple averaging, and combining \textit{Diff} with \textit{Graph} yields the best performance, demonstrating the effectiveness of structured aggregation over multiple reference videos.

\paragraph{\textbf{Branch Configuration.}} We ablate the two branches, where the visual branch captures spatial-temporal features and the alignment branch models video-prompt alignment. As shown in Table~\ref{tab:ablation_all} (d), using only a single branch leads to limited performance. In addition, the alignment branch performs better than the visual one, which may be attributed to the BLIP that takes both visual and textual inputs. When combining both branches, the model achieves the best performance, highlighting their complementarity.

\paragraph{\textbf{Training and Inference Strategies.}} We compare three training and inference strategies, including freezing the backbone (\textit{Frozen Backbone}), end-to-end training (\textit{Joint Training}), and our staged scheme (\textit{Ours}). As shown in Table~\ref{tab:ablation_all} (e), \textit{Frozen Backbone} achieves the lowest training and inference cost due to feature caching, but suffers from a notable performance drop (e.g., SRCC: 0.826 $\rightarrow$ 0.781), ranking only 9th among all methods in Table~\ref{tab:T2VQA}. \textit{Joint Training} maintains strong performance at the expense of substantially increased training time and slightly higher inference cost. In contrast, our method achieves comparable performance to \textit{Joint Training} while reducing training overhead and retaining minimal inference time, resulting in a better trade-off between accuracy and efficiency.


\paragraph{\textbf{Hyperparameter Settings.}} Figure~\ref{fig_parameter} presents the sensitivity of the proposed method to the retrieval threshold $\tau$, loss weight $\gamma$, and batch size $B$. From a global perspective, the four evaluation metrics exhibit consistent trends across different hyperparameters. In particular, SRCC, PLCC, and KRCC remain relatively stable, while RMSE shows slightly larger fluctuations. This difference may be attributed to the fact that the objective combines a correlation-based loss and a ranking-aware loss, which naturally leads to stable performance on rank-based and correlation metrics, while allowing slightly larger variations in RMSE. From a parameter-wise perspective, although minor variations can be observed within each hyperparameter range, the overall performance remains stable, indicating that the proposed method is not sensitive to specific hyperparameter choices. Overall, these results demonstrate that the proposed framework is robust to hyperparameter variations and does not require delicate tuning to achieve strong performance.

\subsection{Visualizations}
We provide comprehensive visualizations to analyze the effectiveness of our reference-aware framework from qualitative, predictive, and representation perspectives. 

\begin{figure}
	\centering
	\includegraphics[width=0.475\textwidth]{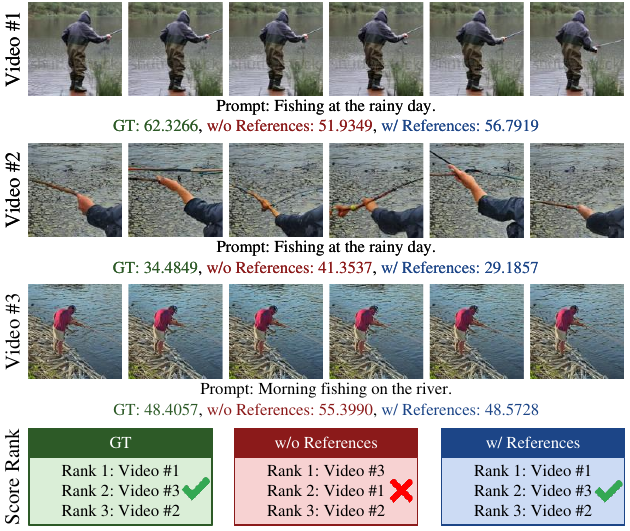}\\
	\vspace{-8pt}
	\caption{Qualitative comparison of quality predictions with and without reference-aware comparative modeling. }
	\vspace{-10pt}
	\label{fig_vis1}
\end{figure}

First, Figure~\ref{fig_vis1} qualitatively demonstrates the effectiveness of reference-aware comparative modeling. Among three videos with similar prompts that are mutually referential, \textit{Video \#2} exhibits incomplete semantics and inconsistent motion compared to \textit{Video \#1}, whereas \textit{Video \#3} remains temporally coherent but appears slightly blurry. By modeling such differences, our method yields more accurate absolute quality predictions and more reliable relative rankings.

\begin{figure}[htbp]
	\centering
	\begin{subfigure}{0.475\textwidth}
		\centering
		\includegraphics[width=\linewidth]{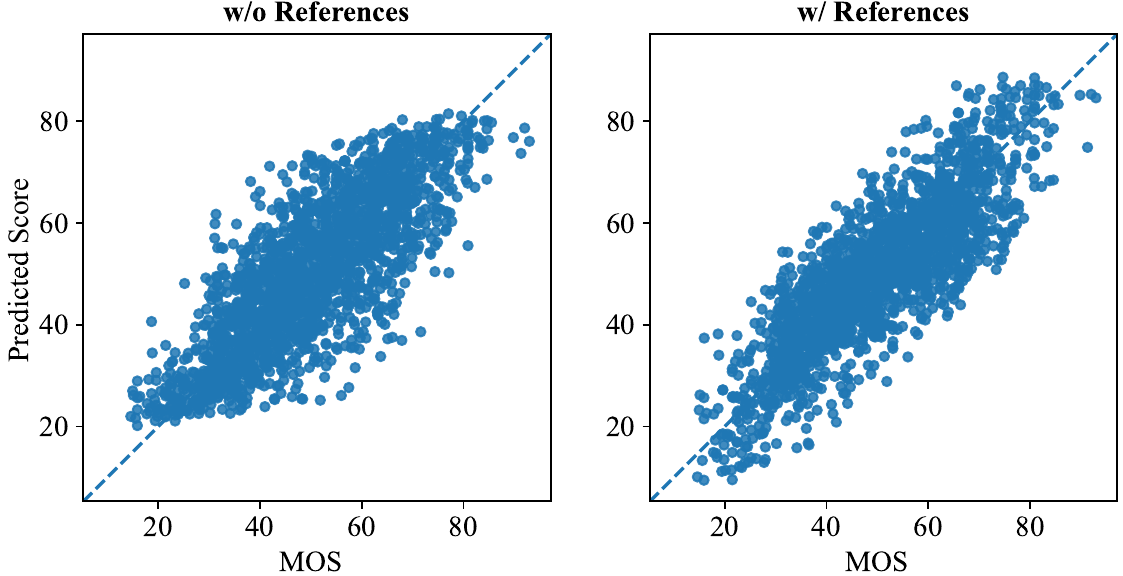}
		\vspace{-12pt}
		\caption{Scatter plots between the predicted score and ground-truth MOS.}
	\end{subfigure}
	\hfill
	\begin{subfigure}{0.475\textwidth}
		\centering
		\includegraphics[width=\linewidth]{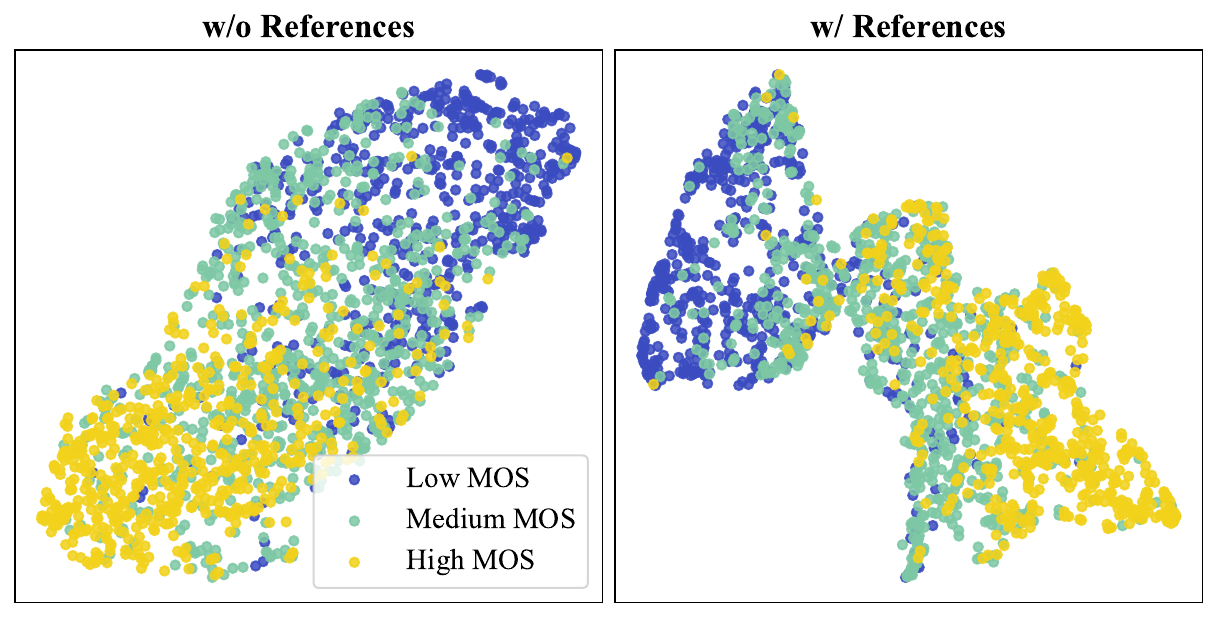}
		\vspace{-10pt}
		\caption{The UMAP visualization of quality features $\mathbf{h}^{q}$, where MOS is uniformly partitioned into low, medium, and high levels.}
	\end{subfigure}
	\vspace{-15pt}
	\caption{Effect of reference-aware comparative modeling on prediction accuracy and feature representations.}
	\vspace{-5pt}
	\label{fig_vis2}
\end{figure}

Second, as shown in Figure~\ref{fig_vis2} (a), incorporating reference information yields predictions that better align with GT MOS. Compared to the reference-free setting, predictions are more concentrated around the diagonal, indicating improved consistency with human judgments. This gain stems from reference-based comparisons that reduce ambiguity in absolute score estimation.

Finally, the Uniform Manifold Approximation and Projection (UMAP) \cite{mcinnes2018umap} visualization in Figure~\ref{fig_vis2}(b) illustrates the clustering of video quality features in a low-dimensional space. The results indicate that without reference-aware comparative modeling, samples from different quality levels largely overlap. When reference information is incorporated, low- and high-quality videos become more clearly separated, forming a discriminative feature space.

\section{Limitations and Future Work}
From the model design perspective, we list several limitations of the proposed method and discuss potential directions for improvement along three key aspects.

\paragraph{\textbf{Reference Video Retrieval.}} Our method retrieves reference videos based on the similarity between prompts. While prompts describe visual factors such as objects, actions, and style, quality assessment focuses on quality factors (e.g., sharpness, artifacts, motion coherence, and semantic consistency). Although perceptual quality is manifested through visual content, it is not directly tied to types of visual factors, and generated videos may also deviate from their prompts, leading to misleading references. Incorporating quality-aware or multimodal retrieval may further improve reference selection.

\paragraph{\textbf{Reference Feature Modeling.}} Our framework adopts a query-centered graph structure, where each reference video is modeled independently with respect to the query. While this design is intuitive and effective, it does not explicitly model relationships among references, including relative quality ordering and shared distortion patterns. More expressive relational modeling strategies, such as full graph propagation, pairwise ranking-based graphs, or set-level attention mechanisms, could further enhance the utilization of interactions within the reference set.

\paragraph{\textbf{Training Paradigm.}} The current training relies on a query-reference paradigm, where each sample is evaluated with retrieved reference videos. This design may lead to repeated use of the same references across different samples, introducing unnecessary computational overhead. A potential alternative is to organize training samples into subsets based on some characteristics, and perform joint evaluation within each subset, allowing shared utilization of reference information. Such a strategy could reduce redundant computations, improve training efficiency, and provide a more effective way to exploit comparative signals during optimization.


\section{Conclusion}

In this paper, we revisit AIGC-VQA from an inter-video perspective and formulate it as a reference-aware evaluation problem, where video quality is assessed in the context of semantically related references. To validate this formulation, we propose the RefVQA model as an effective instantiation, which captures relative visual perceptual and semantic consistency differences via graph-guided difference aggregation. Our method achieves state-of-the-art performance across multiple datasets, quality dimensions, and evaluation settings. More importantly, experimental results demonstrate the effectiveness of incorporating inter-video relationships into feature representations, rather than modeling each sample in isolation. We believe this work will encourage further exploration of reference-aware AIGC-VQA, including improved reference selection, more expressive relational modeling, and more flexible evaluation paradigms.

\appendix

\section{Appendix}
\label{sec:ar}

\subsection{Reference Analysis}
\label{sec:ra}

Table~\ref{rfm} evaluates different reference feature aggregation strategies on T2VQA-DB, including \textit{Average}, \textit{Concatenation}, \textit{Cross-Attention}, and our graph-based method. Our method achieves the best performance across all metrics. Compared to \textit{Average}, it improves SRCC/PLCC/KRCC from 0.821/0.826/0.633 to 0.826/0.835/0.642, indicating more effective use of reference information. This suggests that explicitly modeling the relationships among retrieved references via a similarity-based graph structure enables more effective aggregation of quality cues, compared to uniform or unstructured adaptive aggregation methods.

\begin{table}[htbp]
	\centering
	\caption{Comparison of reference feature aggregation strategies.}
	\footnotesize
	\begin{tabular}{ccccc}
		\toprule
		\multirow{2}{*}{\textbf{Method}} & \multicolumn{4}{c}{\textbf{T2VQA-DB}}\\
		\cmidrule(lr){2-5}
		& SRCC $\uparrow$ & PLCC $\uparrow$ & KRCC $\uparrow$ & RMSE $\downarrow$\\
		\cmidrule{1-5}
		Average & \underline{0.821} & \underline{0.826} & \underline{0.633} & \underline{8.458}\\
		Concatenation & 0.817 & 0.818 & 0.625 & 8.744\\
		Cross-Attention & 0.818 & 0.824 & 0.629 & 8.648\\
		Graph (Ours) & \textbf{0.826} & \textbf{0.835} & \textbf{0.642} & \textbf{8.429}\\
		\bottomrule
	\end{tabular}
	\label{rfm}%
\end{table}%

Table~\ref{rs} analyzes the effect of the similarity threshold $\tau$ on both the number of retrieved references and the final performance. As expected, a smaller $\tau$ leads to a larger reference set, while a larger $\tau$ significantly reduces its size. Notably, non-empty reference sets are consistently obtained across these thresholds. This may be attributed to the intrinsic properties of the dataset, where prompts are reused or exhibit high semantic similarity, increasing the likelihood of retrieving related samples.

Despite the substantial variation in reference number, the performance remains relatively stable. Even with many references, no obvious degradation is observed. We attribute this robustness to our similarity-based aggregation design, which weights reference features based on their similarities, thereby suppressing less informative samples. From a cost-performance perspective, $\tau=0.7$ provides a favorable trade-off, achieving the best performance with a moderate number of references. These results indicate that our method is robust to variations in reference number.

\begin{table}[htbp]
	\centering
	\caption{Effect of the similarity threshold $\tau$ on reference number and performance.}
	\resizebox{0.48\textwidth}{!}{
		\begin{tabular}{cccccccc}
			\toprule
			\multirow{2}{*}{\textbf{Threshold $\tau$}} & \multicolumn{3}{c}{\textbf{Reference Number}} & \multicolumn{4}{c}{\textbf{T2VQA-DB}}\\
			\cmidrule(lr){2-4} 	\cmidrule(lr){5-8}
			& MIN & MAX & AVG & SRCC $\uparrow$ & PLCC $\uparrow$ & KRCC $\uparrow$ & RMSE $\downarrow$\\
			\cmidrule{1-8}
			0.3 & 17 & 2310 & 889.249 & 0.819 &0.823 &0.630 & 8.625 \\
			0.5 & 5 & 897 & 129.551 & 0.816 &0.820 &0.626 &8.680 \\
			0.6 & 4 & 371 & 43.912 & 0.820 &0.825 &0.627 &\underline{8.542} \\
			0.7 & 4 & 149 & 13.871 & \textbf{0.826} & \textbf{0.835} & \textbf{0.642} & \textbf{8.429}\\
			0.8 & 3 & 51 & 8.361 & \underline{0.822} & \underline{0.830} & \underline{0.635} & 8.566 \\
			\bottomrule
		\end{tabular}
	}
	\label{rs}%
\end{table}%

\subsection{Inference Time Analysis}
\label{sec:ita}

Table~\ref{tc} compares inference efficiency and prediction performance across AIGC-VQA methods. We include T2VQA and AIGVEval as baselines since they are both recent and publicly available, enabling a fair and reproducible comparison under consistent settings. RefVQA achieves the best performance on all metrics while requiring the least inference time. Specifically, it improves SRCC/PLCC/KRCC to 0.826/0.835/0.642, compared to 0.797/0.807/0.606 of T2VQA and 0.749/0.764/0.561 of AIGVEval, indicating stronger alignment with human judgments. At the same time, RefVQA reduces RMSE to 8.429 and inference time to 99.265 ms, yielding a clear advantage over both baselines. These results suggest that RefVQA provides a better accuracy-efficiency trade-off.

\begin{table}[htbp]
	\centering
	\caption{Comparison of inference time across different methods. The reported time denotes the average inference time per sample measured on an NVIDIA RTX 4090 GPU.}
	\resizebox{0.48\textwidth}{!}{
		\begin{tabular}{ccccccc}
			\toprule
			\multirow{2}{*}{\textbf{Method}} & \multirow{2}{*}{\textbf{Year}} & \multicolumn{5}{c}{\textbf{T2VQA-DB}}\\
			\cmidrule{3-7}
			& & {Time (ms) $\downarrow$} & {SRCC $\uparrow$} & {PLCC $\uparrow$} & {KRCC $\uparrow$} & {RMSE $\downarrow$}\\
			\cmidrule{1-7}
			T2VQA \cite{kou2024subjective} & ACM MM 24 & \underline{122.187}  & \underline{0.797} & \underline{0.807} & \underline{0.606} & \underline{9.022}\\
			AIGVEval \cite{qi2025towards} & CVPRW 25 & 149.586  & 0.749 & 0.764 & 0.561 & 10.034\\
			\midrule
			RefVQA (Ours) & - & \textbf{99.265} & \textbf{0.826} & \textbf{0.835} & \textbf{0.642} & \textbf{8.429}\\
			\bottomrule
		\end{tabular}%
		\label{tc}%
	}
\end{table}%

\subsection{Ablation Studies on LGVQ}
\label{sec:asl}

\begin{table*}[htbp]
	\centering
	\caption{Ablation study of reference-aware comparative modeling.}
	\footnotesize
	\begin{tabular}{ccccccccccc}
		\toprule
		& & \multicolumn{9}{c}{\textbf{LGVQ}}\\
		\cmidrule(lr){3-11}
		\multicolumn{2}{c}{\textbf{Reference Usage}} & \multicolumn{3}{c}{{Spatial}} & \multicolumn{3}{c}{{Temporal}} & \multicolumn{3}{c}{{Alignment}}\\
		\cmidrule(lr){1-2} \cmidrule(lr){3-5} \cmidrule(lr){6-8} \cmidrule(lr){9-11}
		Visual & Alignment & SRCC $\uparrow$ & PLCC $\uparrow$ & KRCC $\uparrow$ & SRCC $\uparrow$ & PLCC $\uparrow$ & KRCC $\uparrow$ & SRCC $\uparrow$ & PLCC $\uparrow$ & KRCC $\uparrow$ \\
		\cmidrule{1-11}
		& & 0.743$\pm$0.018 & 0.760$\pm$0.014 & 0.551$\pm$0.016 & 0.880$\pm$0.005 & 0.895$\pm$0.004 & 0.686$\pm$0.004 & 0.711$\pm$0.017 & 0.722$\pm$0.010 & 0.528$\pm$0.013 \\
		\checkmark & & \underline{0.762$\pm$0.012} & \underline{0.788$\pm$0.007} & \underline{0.567$\pm$0.009} & \underline{0.897$\pm$0.004} & \underline{0.916$\pm$0.005} & \underline{0.715$\pm$0.007} & 0.717$\pm$0.013 & 0.728$\pm$0.008 & 0.532$\pm$0.010\\
		& \checkmark & 0.745$\pm$0.016 & 0.766$\pm$0.015 & 0.553$\pm$0.014 & 0.884$\pm$0.005 & 0.908$\pm$0.006 & 0.697$\pm$0.006 & \underline{0.727$\pm$0.016} & \underline{0.735$\pm$0.010} & \underline{0.540$\pm$0.011}\\
		\checkmark& \checkmark & \textbf{0.764$\pm$0.013} & \textbf{0.803$\pm$0.007} & \textbf{0.576$\pm$0.011} & \textbf{0.906$\pm$0.004} & \textbf{0.924$\pm$0.004} & \textbf{0.723$\pm$0.005} & \textbf{0.731$\pm$0.014} & \textbf{0.738$\pm$0.011} & \textbf{0.544$\pm$0.009}\\
		\bottomrule
	\end{tabular}%
	\label{as_lgvq}%
\end{table*}%

\begin{figure*}[htbp]
	\centering
	\includegraphics[width=0.95\textwidth]{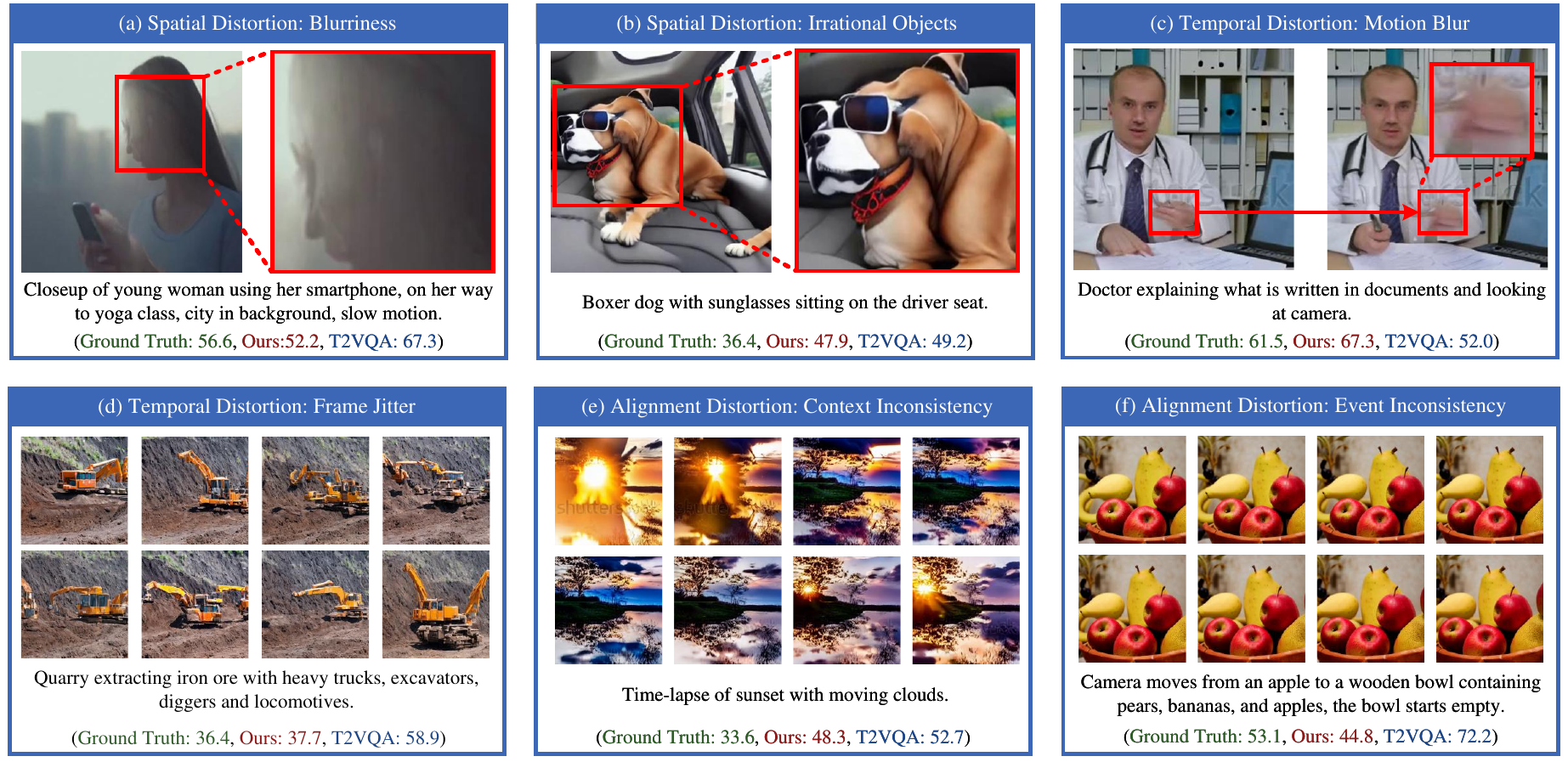}
	\caption{Qualitative comparison of predicted quality scores under different distortion types.}
	\label{fig:vis3}
\end{figure*}

Table~\ref{as_lgvq} presents an ablation study of reference-aware comparative modeling across spatial, temporal, and alignment dimensions on LGVQ. Incorporating either visual or alignment references consistently improves performance over the baseline across all metrics, while combining both yields the best results. The standard deviations remain low across all settings, indicating stable model behavior. Introducing references does not increase variance; instead, it slightly reduces the standard deviation on average compared to the baseline. These results support the effectiveness and reliability of the proposed reference modeling method.

\subsection{Qualitative Comparison}
\label{sec:qc}

Figure~\ref{fig:vis3} presents a qualitative comparison between the proposed method and T2VQA under six representative distortion types. For spatial distortions, our method better captures local quality degradation such as blurriness and unrealistic object structures. In (a), it assigns a lower score to blurred content, while T2VQA overestimates the quality. For temporal distortions, our method is more sensitive to motion-related artifacts. In (d), it avoids the severe overestimation observed in T2VQA under frame jitter, reflecting improved temporal consistency modeling. For alignment distortions, the advantage is more pronounced. Our method better captures both context and event inconsistencies, and its predictions are generally closer to the ground truth. This suggests that reference-aware comparative modeling better reflects semantic alignment between generated videos and their prompts.

Overall, the proposed method produces more reasonable predictions than T2VQA across diverse distortion types. However, accurate score prediction remains challenging in difficult cases, such as irrational objects and context inconsistency. Developing and incorporating distortion-specific backbones for more fine-grained feature extraction could help address these challenges.

\bibliographystyle{ACM-Reference-Format}
\balance
\bibliography{sample-base}

@String{Computing = "Computing" }

@String{Computer = "{IEEE} Computer" }

@String{Springer = "Springer-Verlag" }

@inproceedings{radford2021learning,
  title={Learning Transferable Visual Models from Natural Language Supervision},
  author={Radford, Alec and Kim, Jong Wook and Hallacy, Chris and Ramesh, Aditya and Goh, Gabriel and Agarwal, Sandhini and Sastry, Girish and Askell, Amanda and Mishkin, Pamela and Clark, Jack and others},
  booktitle={International Conference on Machine Learning},
  pages={8748--8763},
  year={2021},
}

@inproceedings{li2022blip,
  title={{BLIP}: Bootstrapping Language-Image Pre-training for Unified Vision-Language Understanding and Generation},
  author={Li, Junnan and Li, Dongxu and Xiong, Caiming and Hoi, Steven},
  booktitle={International Conference on Machine Learning},
  pages={12888--12900},
  year={2022},
}

@inproceedings{reimers2019sentence,
  title={{Sentence-BERT}: Sentence Embeddings using Siamese {BERT}-Networks},
  author={Reimers, Nils and Gurevych, Iryna},
  booktitle={Proceedings of the 2019 Conference on Empirical Methods in Natural Language Processing and the 9th International Joint Conference on Natural Language Processing},
  pages={3982--3992},
  year={2019}
}

@inproceedings{pan2020star,
  title={Star Graph Neural Networks for Session-based Recommendation},
  author={Pan, Zhiqiang and Cai, Fei and Chen, Wanyu and Chen, Honghui and De Rijke, Maarten},
  booktitle={Proceedings of the 29th ACM international Conference on Information \& Knowledge Management},
  pages={1195--1204},
  year={2020}
}

@article{zou2026mds,
  title={{MDS-VQA}: Model-Informed Data Selection for Video Quality Assessment},
  author={Zou, Jian and Xu, Xiaoyu and Wang, Zhihua and Wang, Yilin and Adsumilli, Balu and Ma, Kede},
  journal={arXiv preprint arXiv:2603.11525},
  pages={1--13},
  year={2026}
}

@inproceedings{wang2023internvid,
  title={{InternVid}: A Large-Scale Video-Text Dataset for Multimodal Understanding and Generation},
  author={Wang, Yi and He, Yinan and Li, Yizhuo and Li, Kunchang and Yu, Jiashuo and Ma, Xin and Li, Xinhao and Chen, Guo and Chen, Xinyuan and Wang, Yaohui and others},
  booktitle={International Conference on Learning Representations},
  pages={1--25},
  year={2024}
}

@article{liu2023fetv,
  title={{FETV}: A Benchmark for Fine-Grained Evaluation of Open-Domain Text-to-Video Generation},
  author={Liu, Yuanxin and Li, Lei and Ren, Shuhuai and Gao, Rundong and Li, Shicheng and Chen, Sishuo and Sun, Xu and Hou, Lu},
  journal={Advances in Neural Information Processing Systems},
  volume={36},
  pages={62352--62387},
  year={2023}
}

@inproceedings{sun2022deep,
  title={A Deep Learning based No-reference Quality Assessment Model for {UGC} Videos},
  author={Sun, Wei and Min, Xiongkuo and Lu, Wei and Zhai, Guangtao},
  booktitle={Proceedings of the ACM International Conference on Multimedia},
  pages={856--865},
  year={2022}
}

@article{li2022blindly,
  title={Blindly Assess Quality of In-the-Wild Videos via Quality-Aware Pre-training and Motion Perception},
  author={Li, Bowen and Zhang, Weixia and Tian, Meng and Zhai, Guangtao and Wang, Xianpei},
  journal={IEEE Transactions on Circuits and Systems for Video Technology},
  volume={32},
  number={9},
  pages={5944--5958},
  year={2022},
}

@inproceedings{wu2022fast,
  title={{FAST-VQA}: Efficient End-to-End Video Quality Assessment with Fragment Sampling},
  author={Wu, Haoning and Chen, Chaofeng and Hou, Jingwen and Liao, Liang and Wang, Annan and Sun, Wenxiu and Yan, Qiong and Lin, Weisi},
  booktitle={European Conference on Computer Vision},
  pages={538--554},
  year={2022},
}

@article{wu2022disentangling,
  title={Disentangling Aesthetic and Technical Effects in Video Quality Assessment for User Generated Content},
  author={Wu, Haoning and Liao, Liang and Chen, Chaofeng and Hou, Jingwen and Wang, Annan and Sun, Wenxiu and Yan, Qiong and Lin, Weisi},
  journal={arXiv preprint arXiv:2211.04894},
  pages={1--19},
  year={2022}
}

@article{madhusudana2022image,
  title={Image Quality Assessment Using Contrastive Learning},
  author={Madhusudana, Pavan C and Birkbeck, Neil and Wang, Yilin and Adsumilli, Balu and Bovik, Alan C},
  journal={IEEE Transactions on Image Processing},
  volume={31},
  pages={4149--4161},
  year={2022},
}

@inproceedings{kou2024subjective,
  title={Subjective-Aligned Dataset and Metric for Text-to-Video Quality Assessment},
  author={Kou, Tengchuan and Liu, Xiaohong and Zhang, Zicheng and Li, Chunyi and Wu, Haoning and Min, Xiongkuo and Zhai, Guangtao and Liu, Ning},
  booktitle={Proceedings of the ACM International Conference on Multimedia},
  pages={7793--7802},
  year={2024}
}

@inproceedings{lu2024aigc,
  title={{AIGC-VQA}: A Holistic Perception Metric for {AIGC} Video Quality Assessment},
  author={Lu, Yiting and Li, Xin and Li, Bingchen and Yu, Zihao and Guan, Fengbin and Wang, Xinrui and Liao, Ruling and Ye, Yan and Chen, Zhibo},
  booktitle={Proceedings of the IEEE/CVF Conference on Computer Vision and Pattern Recognition Workshops},
  pages={6384--6394},
  year={2024}
}

@inproceedings{cheng2025vpo,
  title={{VPO}: Aligning Text-to-Video Generation Models with Prompt Optimization},
  author={Cheng, Jiale and Lyu, Ruiliang and Gu, Xiaotao and Liu, Xiao and Xu, Jiazheng and Lu, Yida and Teng, Jiayan and Yang, Zhuoyi and Dong, Yuxiao and Tang, Jie and others},
  booktitle={Proceedings of the IEEE/CVF International Conference on Computer Vision},
  pages={15636--15645},
  year={2025}
}

@inproceedings{qu2024exploring,
  title={Exploring {AIGC} Video Quality: A Focus on Visual Harmony Video-Text Consistency and Domain Distribution Gap},
  author={Qu, Bowen and Liang, Xiaoyu and Sun, Shangkun and Gao, Wei},
  booktitle={Proceedings of the IEEE/CVF Conference on Computer Vision and Pattern Recognition Workshops},
  pages={6652--6660},
  year={2024}
}

@article{mcinnes2018umap,
  title={{UMAP}: Uniform Manifold Approximation and Projection for Dimension Reduction},
  author={McInnes, Leland and Healy, John and Melville, James},
  journal={arXiv preprint arXiv:1802.03426},
  pages={1--63},
  year={2018}
}

@article{min2024perceptual,
  title={Perceptual Video Quality Assessment: A Survey},
  author={Min, Xiongkuo and Duan, Huiyu and Sun, Wei and Zhu, Yucheng and Zhai, Guangtao},
  journal={Science China Information Sciences},
  volume={67},
  number={11},
  pages={211301},
  year={2024},
}

@inproceedings{chen2025vquala,
  title={{VQualA} 2025 Challenge on {GenAI-Bench AIGC} Video Quality Assessment: Methods and Results},
  author={Chen, Ying and Wang, Huasheng and Xiao, Pengxiang and Ding, Yukang and Liu, Enpeng and Zhou, Chris Wei and Li, Baojun and Huang, Jiamian and Wang, Jiarui and Zhu, Xiaorong and others},
  booktitle={IEEE/CVF International Conference on Computer Vision Workshops (ICCVW)},
  pages={3422--3431},
  year={2025},
  organization={IEEE}
}

@inproceedings{qi2025towards,
  title={Towards Holistic Visual Quality Assessment of {AI}-Generated Videos: A {LLM}-Based Multi-Dimensional Evaluation Model},
  author={Qi, Zelu and Shi, Ping and Zhang, Chaoyang and Wang, Shuqi and Zhao, Fei and Pan, Da and Ying, Zefeng},
  booktitle={Proceedings of the IEEE/CVF Conference on Computer Vision and Pattern Recognition Workshops},
  pages={1493--1502},
  year={2025}
}

@inproceedings{zhang2025q,
  title={{Q-Eval-100K}: Evaluating Visual Quality and Alignment Level for Text-to-Vision Content},
  author={Zhang, Zicheng and Kou, Tengchuan and Wang, Shushi and Li, Chunyi and Sun, Wei and Wang, Wei and Li, Xiaoyu and Wang, Zongyu and Cao, Xuezhi and Min, Xiongkuo and others},
  booktitle={Proceedings of the IEEE/CVF Conference on Computer Vision and Pattern Recognition},
  pages={10621--10631},
  year={2025}
}

@article{qi2025t2veval,
  title={{T2VEval}: Benchmark Dataset and Objective Evaluation Method for {T2V}-generated Videos},
  author={Qi, Zelu and Shi, Ping and Wang, Shuqi and Zhang, Chaoyang and Zhao, Fei and Ying, Zefeng and Pan, Da and Yang, Xi and He, Zheqi and Dai, Teng},
  journal={arXiv preprint arXiv:2501.08545},
  pages={1--18},
  year={2025}
}

@article{yue2024advancing,
  title={Advancing Video Quality Assessment for {AIGC}},
  author={Yue, Xinli and Sun, Jianhui and Kong, Han and Yao, Liangchao and Wang, Tianyi and Li, Lei and Rao, Fengyun and Lv, Jing and Xia, Fan and Deng, Yuetang and others},
  journal={arXiv preprint arXiv:2409.14888},
  pages={1--5},
  year={2024}
}

@inproceedings{li2025multilevel,
  title={Multilevel Semantic-Aware Model for {AI}-Generated Video Quality Assessment},
  author={Li, Jiaze and Xu, Haoran and Zhu, Shiding and He, Junwei and Wang, Haozhao},
  booktitle={IEEE International Conference on Acoustics, Speech and Signal Processing},
  pages={1--5},
  year={2025},
}

@article{zhang2025lmvq,
  title={{LMVQ}: Label-free Metric-learning for General {AI}-generated Video Quality Assessment},
  author={Zhang, Zhichao and Li, Xinyue and Sun, Wei and Zhang, Zicheng and Liu, Xiaohong and Min, Xiongkuo and Zhai, Guangtao},
  journal={IEEE Transactions on Circuits and Systems for Video Technology},
note = {Early Access},
  year={2025},
}

@InProceedings{Liu_2022_CVPR,
    author    = {Liu, Ze and Ning, Jia and Cao, Yue and Wei, Yixuan and Zhang, Zheng and Lin, Stephen and Hu, Han},
    title     = {{Video Swin Transformer}},
    booktitle = {Proceedings of the IEEE/CVF Conference on Computer Vision and Pattern Recognition},
    month     = {June},
    year      = {2022},
    pages     = {3202-3211}
}

@inproceedings{wang2025aigv,
  title={{AIGV-Assessor}: Benchmarking and Evaluating the Perceptual Quality of Text-to-Video Generation with {LMM}s},
  author={Wang, Jiarui and Duan, Huiyu and Zhai, Guangtao and Wang, Juntong and Min, Xiongkuo},
  booktitle={Proceedings of the Computer Vision and Pattern Recognition Conference},
  pages={18869--18880},
  year={2025}
}

@article{sun2025content,
  title={Content-Rich {AIGC} Video Quality Assessment via Intricate Text Alignment and Motion-Aware Consistency},
  author={Sun, Shangkun and Liang, Xiaoyu and Qu, Bowen and Gao, Wei},
  journal={arXiv preprint arXiv:2502.04076},
  pages={1--13},
  year={2025}
}

@inproceedings{sun2025ve,
  title={{VE-Bench}: Subjective-Aligned Benchmark Suite for Text-Driven Video Editing Quality Assessment},
  author={Sun, Shangkun and Liang, Xiaoyu and Fan, Songlin and Gao, Wenxu and Gao, Wei},
  booktitle={Proceedings of the AAAI Conference on Artificial Intelligence},
  volume={39},
  number={7},
  pages={7105--7113},
  year={2025}
}

@inproceedings{ke2021musiq,
  title={{MUSIQ}: Multi-Scale Image Quality Transformer},
  author={Ke, Junjie and Wang, Qifei and Wang, Yilin and Milanfar, Peyman and Yang, Feng},
  booktitle={Proceedings of the IEEE/CVF International Conference on Computer Vision},
  pages={5148--5157},
  year={2021}
}

@article{sun2023blind,
  title={Blind Quality Assessment for in-the-Wild Images via Hierarchical Feature Fusion and Iterative Mixed Database Training},
  author={Sun, Wei and Min, Xiongkuo and Tu, Danyang and Ma, Siwei and Zhai, Guangtao},
  journal={IEEE Journal of Selected Topics in Signal Processing},
  volume={17},
  number={6},
  pages={1178--1192},
  year={2023},
}

@inproceedings{wang2023exploring,
  title={Exploring {CLIP} for Assessing the Look and Feel of Images},
  author={Wang, Jianyi and Chan, Kelvin CK and Loy, Chen Change},
  booktitle={Proceedings of the AAAI Conference on Artificial Intelligence},
  volume={37},
  number={2},
  pages={2555--2563},
  year={2023}
}

@inproceedings{zhang2023blind,
  title={Blind Image Quality Assessment via Vision-Language Correspondence: A Multitask Learning Perspective},
  author={Zhang, Weixia and Zhai, Guangtao and Wei, Ying and Yang, Xiaokang and Ma, Kede},
  booktitle={Proceedings of the IEEE/CVF Conference on Computer Vision and Pattern Recognition},
  pages={14071--14081},
  year={2023}
}

@article{zhang2024benchmarking,
  title={Benchmarking Multi-dimensional {AIGC} Video Quality Assessment: A Dataset and Unified Model},
  author={Zhang, Zhichao and Sun, Wei and Xinyue, Li and Jia, Jun and Min, Xiongkuo and Zhang, Zicheng and Li, Chunyi and Chen, Zijian and Puyi, Wang and Fengyu, Sun and others},
  journal={ACM Transactions on Multimedia Computing, Communications and Applications},
  volume={21},
  number={9},
  pages={1--24},
  year={2025},
}

@article{kay2017kinetics,
  title={The Kinetics Human Action Video Dataset},
  author={Kay, Will and Carreira, Joao and Simonyan, Karen and Zhang, Brian and Hillier, Chloe and Vijayanarasimhan, Sudheendra and Viola, Fabio and Green, Tim and Back, Trevor and Natsev, Paul and others},
  journal={arXiv preprint arXiv:1705.06950},
  pages={1--22},
  year={2017}
}

@ARTICLE{9633248,
  author={Kancharla, Parimala and Channappayya, Sumohana S.},
  journal={IEEE Transactions on Image Processing}, 
  title={Completely Blind Quality Assessment of User Generated Video Content}, 
  year={2022},
  volume={31},
  number={},
  pages={263-274},
}

@inproceedings{wu2023exploring,
  title={Exploring Video Quality Assessment on User Generated Contents from Aesthetic and Technical Perspectives},
  author={Wu, Haoning and Zhang, Erli and Liao, Liang and Chen, Chaofeng and Hou, Jingwen and Wang, Annan and Sun, Wenxiu and Yan, Qiong and Lin, Weisi},
  booktitle={Proceedings of the IEEE/CVF International Conference on Computer Vision},
  pages={20144--20154},
  year={2023}
}

@inproceedings{hessel2021clipscore,
  title={{CLIPScore}: A Reference-Free Evaluation Metric for Image Captioning},
  author={Hessel, Jack and Holtzman, Ari and Forbes, Maxwell and Le Bras, Ronan and Choi, Yejin},
  booktitle={Proceedings of the 2021 Conference on Empirical Methods in Natural Language Processing},
  pages={7514--7528},
  year={2021}
}

@article{xu2023imagereward,
title={ImageReward: Learning and evaluating human preferences for text-to-image generation},
  author={Xu, Jiazheng and Liu, Xiao and Wu, Yuchen and Tong, Yuxuan and Li, Qinkai and Ding, Ming and Tang, Jie and Dong, Yuxiao},
  journal={Advances in Neural Information Processing Systems},
  volume={36},
  pages={15903--15935},
  year={2023}
}

@article{kirstain2023pick,
  title={{Pick-a-Pic}: An Open Dataset of User Preferences for Text-to-Image Generation},
  author={Kirstain, Yuval and Polyak, Adam and Singer, Uriel and Matiana, Shahbuland and Penna, Joe and Levy, Omer},
  journal={Advances in Neural Information Processing Systems},
  volume={36},
  pages={36652--36663},
  year={2023}
}

@article{wu2023human,
  title={Human Preference Score v2: A Solid Benchmark for Evaluating Human Preferences of Text-to-Image Synthesis},
  author={Wu, Xiaoshi and Hao, Yiming and Sun, Keqiang and Chen, Yixiong and Zhu, Feng and Zhao, Rui and Li, Hongsheng},
  journal={arXiv preprint arXiv:2306.09341},
  pages={1--24},
  year={2023}
}

@article{zhang2021uncertainty,
  title={Uncertainty-Aware Blind Image Quality Assessment in the Laboratory and Wild},
  author={Zhang, Weixia and Ma, Kede and Zhai, Guangtao and Yang, Xiaokang},
  journal={IEEE Transactions on Image Processing},
  volume={30},
  pages={3474--3486},
  year={2021},
}

@inproceedings{wei2025dreamrelation,
  title={{DreamRelation}: Relation-Centric Video Customization},
  author={Wei, Yujie and Zhang, Shiwei and Yuan, Hangjie and Gong, Biao and Tang, Longxiang and Wang, Xiang and Qiu, Haonan and Li, Hengjia and Tan, Shuai and Zhang, Yingya and others},
  booktitle={Proceedings of the IEEE/CVF International Conference on Computer Vision},
  pages={12381--12393},
  year={2025}
}

@article{zhu2022relational,
  title={Relational Reasoning Over Spatial-Temporal Graphs for Video Summarization},
  author={Zhu, Wencheng and Han, Yucheng and Lu, Jiwen and Zhou, Jie},
  journal={IEEE Transactions on Image Processing},
  volume={31},
  pages={3017--3031},
  year={2022},
  publisher={IEEE}
}

@article{liu2023cross,
  title={Cross-Modal Causal Relational Reasoning for Event-Level Visual Question Answering},
  author={Liu, Yang and Li, Guanbin and Lin, Liang},
  journal={IEEE Transactions on Pattern Analysis and Machine Intelligence},
  volume={45},
  number={10},
  pages={11624--11641},
  year={2023},
  publisher={IEEE}
}

@article{bai2024event,
  title={Event Graph Guided Compositional Spatial-Temporal Reasoning for Video Question Answering},
  author={Bai, Ziyi and Wang, Ruiping and Gao, Difei and Chen, Xilin},
  journal={IEEE Transactions on Image Processing},
  volume={33},
  pages={1109--1121},
  year={2024},
  publisher={IEEE}
}

@article{chen2024survey,
  title={A Survey on Graph Neural Networks and Graph Transformers in Computer Vision: A Task-Oriented Perspective},
  author={Chen, Chaoqi and Wu, Yushuang and Dai, Qiyuan and Zhou, Hong-Yu and Xu, Mutian and Yang, Sibei and Han, Xiaoguang and Yu, Yizhou},
  journal={IEEE Transactions on Pattern Analysis and Machine Intelligence},
  volume={46},
  number={12},
  pages={10297--10318},
  year={2024},
  publisher={IEEE}
}

@inproceedings{xing2023boosting,
  title={Boosting Few-shot Action Recognition with Graph-guided Hybrid Matching},
  author={Xing, Jiazheng and Wang, Mengmeng and Ruan, Yudi and Chen, Bofan and Guo, Yaowei and Mu, Boyu and Dai, Guang and Wang, Jingdong and Liu, Yong},
  booktitle={Proceedings of the IEEE/CVF international conference on computer vision},
  pages={1740--1750},
  year={2023}
}

@article{zhou2025videognn,
  title={{VideoGNN}: Video Representation Learning via Dynamic Graph Modelling},
  author={Zhou, Zijie and Zhou, Mingliang and Luo, Jun and Pu, Huayan and U, Leong Hou and Wei, Xuekai and Jia, Weijia},
  journal={ACM Transactions on Multimedia Computing, Communications and Applications},
  volume={21},
  number={12},
  pages={1--22},
  year={2025},
}

@inproceedings{qiu2025step,
  title={{STEP}: Enhancing {Video-LLMs'} Compositional Reasoning by Spatio-Temporal Graph-guided Self-Training},
  author={Qiu, Haiyi and Gao, Minghe and Qian, Long and Pan, Kaihang and Yu, Qifan and Li, Juncheng and Wang, Wenjie and Tang, Siliang and Zhuang, Yueting and Chua, Tat-Seng},
  booktitle={Proceedings of the IEEE/CVF Conference on Computer Vision and Pattern Recognition},
  pages={3284--3294},
  year={2025}
}

@inproceedings{wang2023vqa,
  title={VQA-GNN: Reasoning with Multimodal Knowledge via Graph Neural Networks for Visual Question Answering},
  author={Wang, Yanan and Yasunaga, Michihiro and Ren, Hongyu and Wada, Shinya and Leskovec, Jure},
  booktitle={Proceedings of the IEEE/CVF International Conference on Computer Vision},
  pages={21582--21592},
  year={2023}
}

@article{you2020graph,
  title={Graph Contrastive Learning with Augmentations},
  author={You, Yuning and Chen, Tianlong and Sui, Yongduo and Chen, Ting and Wang, Zhangyang and Shen, Yang},
  journal={Advances in Neural Information Processing Systems},
  volume={33},
  pages={5812--5823},
  year={2020}
}

@inproceedings{zeng2021multi,
  title={Multi-Modal Relational Graph for Cross-Modal Video Moment Retrieval},
  author={Zeng, Yawen and Cao, Da and Wei, Xiaochi and Liu, Meng and Zhao, Zhou and Qin, Zheng},
  booktitle={Proceedings of the IEEE/CVF Conference on Computer Vision and Pattern Recognition},
  pages={2215--2224},
  year={2021}
}

@article{sang2020context,
  title={Context-Dependent Propagating-Based Video Recommendation in Multimodal Heterogeneous Information Networks},
  author={Sang, Lei and Xu, Min and Qian, Shengsheng and Martin, Matt and Li, Peter and Wu, Xindong},
  journal={IEEE Transactions on Multimedia},
  volume={23},
  pages={2019--2032},
  year={2020},
  publisher={IEEE}
}

@inproceedings{xiao2022video,
  title={Video Graph Transformer for Video Question Answering},
  author={Xiao, Junbin and Zhou, Pan and Chua, Tat-Seng and Yan, Shuicheng},
  booktitle={European Conference on Computer Vision},
  pages={39--58},
  year={2022},
  organization={Springer}
}

@inproceedings{gao2023generalized,
  title={Generalized Relation Modeling for Transformer Tracking},
  author={Gao, Shenyuan and Zhou, Chunluan and Zhang, Jun},
  booktitle={Proceedings of the IEEE/CVF Conference on Computer Vision and Pattern Recognition},
  pages={18686--18695},
  year={2023}
}

@article{zhang2025vq,
  title={{VQ-Insight}: Teaching {VLM}s for {AI}-Generated Video Quality Understanding via Progressive Visual Reinforcement Learning},
  author={Zhang, Xuanyu and Li, Weiqi and Zhao, Shijie and Li, Junlin and Zhang, Li and Zhang, Jian},
  booktitle={Proceedings of the AAAI Conference on Artificial Intelligence},
  volume={40},
  number={15},
  pages={12870--12878},
  year={2026}
}

@ARTICLE{9405420,
  author={Tu, Zhengzhong and Wang, Yilin and Birkbeck, Neil and Adsumilli, Balu and Bovik, Alan C.},
  journal={IEEE Transactions on Image Processing}, 
  title={{UGC-VQA}: Benchmarking Blind Video Quality Assessment for User Generated Content}, 
  year={2021},
  volume={30},
  pages={4449-4464},
}

\end{document}